%% file: FSOD-CVPR23.tex
\documentclass[10pt,twocolumn,letterpaper]{article}

\usepackage[pagenumbers]{cvpr}

\usepackage[utf8]{inputenc} 
\usepackage[T1]{fontenc}    
\usepackage{url}            
\usepackage{booktabs}       
\usepackage{amsfonts}       
\usepackage{nicefrac}       
\usepackage{microtype}      
\usepackage{xcolor}         

\usepackage{graphicx}
\usepackage{subcaption}

\def\HangBox#1{
\begin{minipage}[t]{\textwidth}
\begin{tabbing} 
~\\[-\baselineskip] 
#1 
\end{tabbing}
\end{minipage}} 

\usepackage{algorithm}
\usepackage[noend]{algpseudocode}

\usepackage{amsmath}
\usepackage{amssymb}
\usepackage{mathtools}
\usepackage{amsthm}
\usepackage{multirow}

\input{macros}

\usepackage{pifont}
\newcommand{\cmark}{\ding{51}}%
\newcommand{\xmark}{\ding{55}}%

\usepackage[pagebackref,breaklinks,colorlinks]{hyperref}

\usepackage[capitalize]{cleveref}
\crefname{section}{Sec.}{Secs.}
\Crefname{section}{Section}{Sections}
\Crefname{table}{Table}{Tables}
\crefname{table}{Tab.}{Tabs.}

\begin{document}

\title{Meta-tuning Loss Functions and Data Augmentation for Few-shot Object Detection}

\author{Berkan Demirel$^{1,2}$ ~\quad Orhun Buğra Baran$^1$ ~\quad Ramazan Gokberk Cinbis$^1$ \quad 
\\
$^1$Middle East Technical University ~\quad 
$^2$HAVELSAN Inc. \\
{\tt\small bdemirel@havelsan.com.tr~\quad bugra@ceng.metu.edu.tr~\quad gcinbis@ceng.metu.edu.tr}
}

\input{oursymbols}

\maketitle

\begin{abstract}
\input{abstract}
\end{abstract}

\input{intro}

\input{relwork}

\input{method}

\input{experiments}

\input{conclusion}

\mypar{Acknowledgements} This work was supported in part by the TUBITAK Grant 119E597 and a Google Faculty Research Award.


{\small
\bibliographystyle{ieee_fullname}
\bibliography{egbib}
}

\newpage

\appendix

\input{appendix}

\end{document}

%% file: macros.tex
\usepackage{color}
\usepackage[normalem]{ulem}

\usepackage{xcolor}
\usepackage{soul}
\usepackage[normalem]{ulem}

\def\mypar#1{\vspace{0.15cm}\noindent{\bf #1.}}

\def\todo#1{{\color{blue}{}}}

\IfFileExists{darkmode.tex}{
    \usepackage{pagecolor}
    \definecolor{myfg}{gray}{0.95} 
    \definecolor{mybg}{gray}{0.2}
    \input{darkmode.tex} 
    \pagecolor{mybg}
    \color{myfg}
}{} 

\usepackage{amssymb} 

\def\etal{et al.\ }

\usepackage{amsmath,amssymb,amsbsy,xspace}
\makeatletter
\DeclareRobustCommand\onedot{\futurelet\@let@token\@onedot}
\def\@onedot{\ifx\@let@token.\else.\null\fi\xspace}

\def\eg{\emph{e.g}\onedot} 
\def\ie{\emph{i.e}\onedot}

\def\etal{\emph{et al}\onedot}

\makeatother

%% file: oursymbols.tex
\def\Cpxb{C_\text{p-base}}
\def\Cpxn{C_\text{p-novel}} 
\def\Dpxpretrain{D_\text{p-pretrain}}
\def\Dpxsupport{D_\text{p-support}} 
\def\Dpxquery{D_\text{p-query}} 
\def\etal{\textit{et al.}}

%% file: abstract.tex
Few-shot object detection, the problem of modelling novel object detection categories with few training instances, is an emerging topic in the area of few-shot learning and object detection. Contemporary techniques can be divided into two groups: fine-tuning based and meta-learning based approaches. While meta-learning approaches aim to learn dedicated meta-models for mapping samples to novel class models, fine-tuning approaches tackle few-shot detection in a simpler manner, by adapting the detection model to novel classes through gradient based optimization. Despite their simplicity, fine-tuning based approaches typically yield competitive detection results. Based on this observation, we focus on the role of loss functions and augmentations as the force driving the fine-tuning process, and propose to tune their dynamics through meta-learning principles. The proposed training scheme, therefore, allows learning inductive biases that can boost few-shot detection, while keeping the advantages of fine-tuning based approaches. In addition, the proposed approach yields interpretable loss functions, as opposed to highly parametric and complex few-shot meta-models. The experimental results highlight the merits of the proposed scheme, with significant improvements over the strong fine-tuning based few-shot detection baselines on benchmark Pascal VOC and MS-COCO datasets, in terms of both standard and generalized few-shot performance metrics.

%% file: intro.tex
\section{Introduction}

\input{fig_intro.tex}

Object detection is one of the computer vision problems that has greatly benefited from the advances
in supervised deep learning approaches. However, similar to the case in many other problems, state-of-the-art in object detection relies on the availability of large-scale fully-annotated datasets,
which is particularly problematic due to the difficulty of collecting
accurate bounding box annotations~\cite{liu2021Swin, Ghiasi2021SimpleCI}. This practical burden has lead to a great interest in
the approaches that can potentially reduce the annotation cost, such as weakly-supervised
learning~\cite{ren-cvpr020, huang2020comprehensive}, learning from point annotations~\cite{Chen2021PointsAQ}, and mixed supervised learning~\cite{liu2021mixed}. A
more recently emerging paradigm in this direction 
is {\em few-shot object detection} (FSOD).
In the FSOD problem, the goal is to build detection models for the {\em novel} classes with few
labeled training images by transferring knowledge from the {\em base} classes with a large set of
training images.  In the closely related Generalized-FSOD (G-FSOD) problem, the goal is to build few-shot detection models that perform
well on both base and novel classes.

FSOD methods can be categorized into meta-learning and fine-tuning approaches.  Although
meta-learning based methods are predominantly used in the literature in FSOD
research~\cite{xiao2020few, zhang2021meta, li2021beyond, chen2021dual, yan2019meta, kang2019few,
perez2020incremental, yin2022sylph, zhang2022kernelized, han2022few}, several fine-tuning based works have recently reported competitive
results~\cite{wang2020frustratingly, sun2021fsce, wu2020multi, fan2021generalized, zhang2021hallucination, cao2021few, kaul2022label, qiao2021defrcn}. The main
premise of meta-learning approaches is 
to design and train dedicated meta-models that map given few train samples to novel class detection
models, \eg~\cite{wu2020meta} or learn easy-to-adapt models~\cite{ood_maml_neurips_2020} 
in a MAML~\cite{finn2017maml} fashion.
In contrast, however, fine-tuning based methods tackle the problem as a typical transfer learning problem
and apply the general purpose supervised training techniques, \ie regularized loss minimization via
gradient-based optimization, to adapt a pre-trained model to few-shot classes.  It is also worth noting that the
recent results on fine-tuning based FSOD are aligned with related observations on few-shot
classification~\cite{tian_rethinking_2020,dhillon_baseline_2020,chen_closer_2019} and segmentation~\cite{boudiaf_few_shot_2021}.

While some of the FSOD meta-learning approaches are attractive for being able to learn dedicated
parametric training mechanisms, they also come with two important shortcomings: (i)
the risk of overfitting to the base classes used for training the meta-model due to model
complexity,
and (ii) the difficulty of interpreting what is actually learned; both of which can be crucially
important for real-world, in-the-wild utilization of a meta-learned model.  From this point of view,
the simplicity and generality of a fine-tuning based FSOD approach can be seen as major advantages. In
fact, one can find a large machine learning literature on the components (optimization techniques,
loss functions, data augmentation, and architectures) of an FT approach, as opposed to the unique and typically unknown nature of
a meta-learned inference model, especially when the model aims to replace standard training
procedures for modeling the novel few-shot classes. While MAML~\cite{finn2017maml} like meta-learning for quick adaptation is
closer in nature to fine-tuning based approaches, the vanishing gradient problems and
the overall complexity of the meta-learning task practically limits the approach to target only one or few model update steps, whereas an FT approach has no such computational difficulty.

Perhaps the biggest advantage of a fine-tuning based FSOD approach, however, can also be its biggest
disadvantage: its generality may lack the inductive biases needed for effective learning with few novel class
samples while preserving the knowledge of base classes.
To this end, such approaches focus on the design of
fine-tuning details, \eg whether to freeze the representation parameters~\cite{wang2020frustratingly}, use contrastive fine-tuning
losses~\cite{sun2021fsce}, increase the novel class variances~\cite{zhang2021hallucination}, introduce the using additional detection heads and branches~\cite{wu2020multi, fan2021generalized}. 
However, optimizing such details specifically for few-shot classes in a hand-crafted manner is
clearly difficult, and likely to be sub-optimal.

To address this problem, we focus on applying meta-learning principles to tune the loss functions and augmentations to be used in the fine-tuning stage for FSOD, which we call {\em meta-tuning} (Figure~\ref{fig:intro}). More specifically, much like the
meta-learning of a meta-model, we define an episodic training procedure that aims to progressively
discover the optimal loss function and augmentation details for FSOD purposes in a data-driven manner.
Using reinforcement learning (RL) techniques, we aim to tune the loss function and augmentation details such that 
they maximize the expected detection quality of an FSOD model obtained by fine-tuning to a set of novel classes. By defining meta-tuning over well-designed loss terms and an augmentation list, we restrict the search process to effective function families, reducing the computational costs compared to 
AutoML methods that aim to discover loss terms from scratch for fully-supervised learning~\cite{liu2021loss, gonzalez2020improved}. 
The resulting meta-tuned loss functions and augmentations, therefore, inject the learned FSOD-specific inductive biases into a fine-tuning based approach. 

To explore the potential of the meta-tuning scheme for FSOD, we focus on the details of classification loss functions,
based on the observations that
FSOD prediction mistakes tend to be in classification rather than localization details~\cite{sun2021fsce}.
In particular, we first focus on the softmax temperature parameter, for which we define two versions: (i) a simple constant temperature, and (ii) time (fine-tuning iteration index) varying dynamic temperature, parameterized as an exponentiated polynomial. In all cases, the parameters learned via meta-tuning yield an interpretable loss function that has a negligible risk of over-fitting to the base classes, in contrast to a complex meta-model. We also model augmentation magnitudes during meta-tuning for improving the data loading pipeline for few-shot learning purposes. Additionally, we incorporate a score scaling coefficient for learning to balance base versus novel class scores.

We provide an experimental analysis on the Pascal VOC~\cite{everingham2010pascal} and
MS-COCO~\cite{lin2014microsoft} benchmarks for FSOD, using the state-of-the-art fine-tuning based
baselines MPSR~\cite{wu2020multi} and DeFRCN~\cite{qiao2021defrcn}.  Our experimental results
show that the proposed meta-tuning approach provides significant performance gains in both FSOD and
Generalized FSOD settings, suggesting that meta-tuning loss functions and data augmentation can be a promising direction in FSOD research.

%% file: fig_intro.tex
\begin{figure}
\vskip -0.2in
\centering
\centerline{\includegraphics[width=\linewidth]{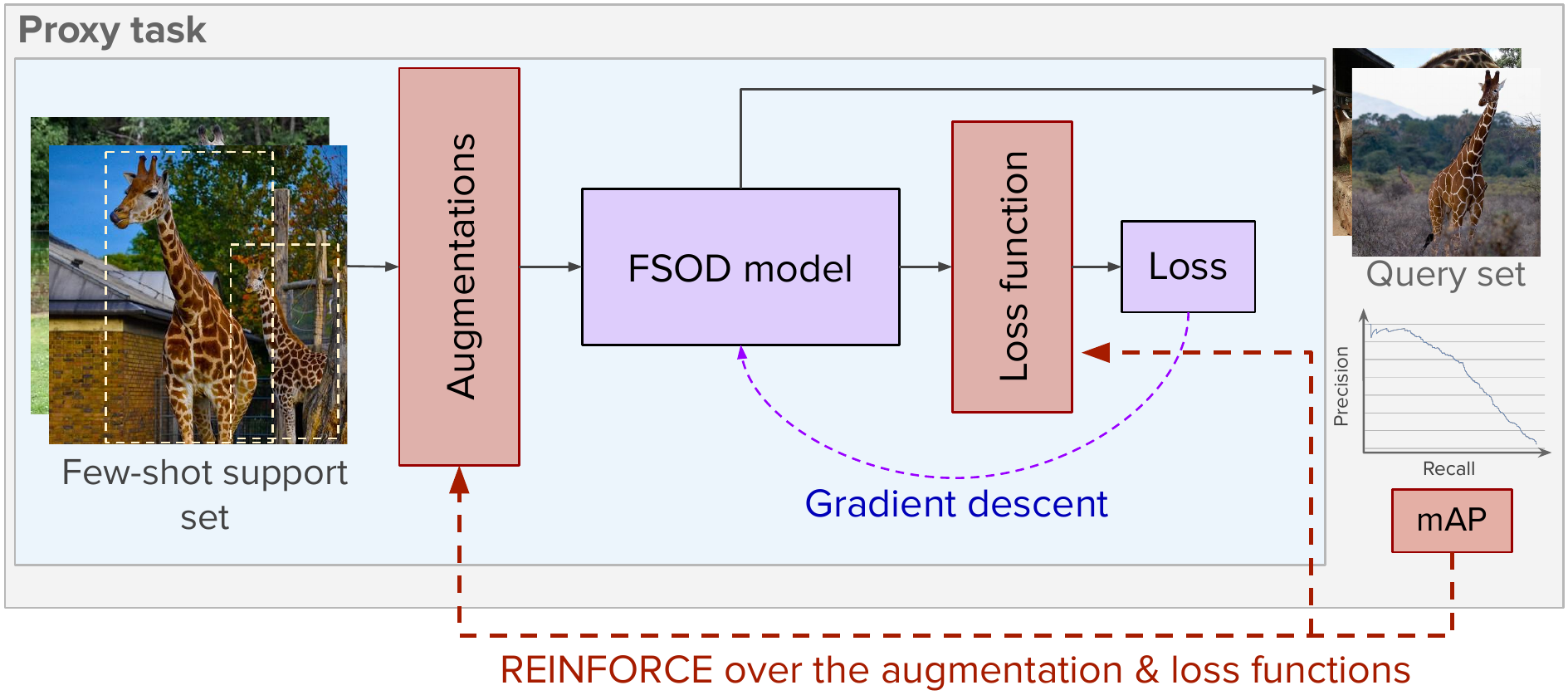}}
    \caption{The overall architecture of the meta-tuning approach.}  \label{fig:intro}
\vskip -0.2in
\end{figure}

%% file: relwork.tex
\section{Related Work}

This section provides an overview of recent developments on few-shot image classification, few-shot
object detection, automated loss function and data augmentation discovery.

\mypar{Few-shot classification} Most of the meta-learning approaches for few-shot learning (FSL) of
classification models can be grouped as {\em adaptation-based} and {\em mapping-based} approaches.
Adaptation-based (also called {\em gradient-based}) approaches aim to learn model parameters that
can easily be adapted to new unseen few-shot tasks within a few model update steps, \eg \cite{finn2017,
li2017, nichol2018, Rajeswaran2019, rusu2018, Munkhdalai2018RapidAW, Park2019MetaCurvature}.
Mapping-based approaches (also called {\em metric-based}) aim to bypass a gradient-descent
based adaptation step, and instead learn a
data-to-classifier mapping, \eg \cite{cao21, liu2019fewTPN, snell2017, sung2017,santoro2016,
vinyals2016, Ye2020FewShotLV, tadam, Zhang_2020_CVPR, mlti2022}.

Some of the other notable approaches include learning to generate synthetic data for novel
classes~\cite{Hariharan2017LowShotVR, wang2018, Lazarou2021TensorFH}, using better feature
representations \cite{tian_rethinking_2020, gidaris2019boosting, wang2019simpleshot, vitfsl2022, cnaps2021, feti2022, negmargin2020} 
or utilizing differentiable convex solvers \cite{bertinetto2018metalearning,
Lee2019MetaLearningWD}.  Importantly, several works highlight that a carefully trained
representation combined with simple fine-tuning or even just shallow classifiers can yield
competitive or better performance than meta-learning based approaches, \eg
\cite{tian_rethinking_2020,dhillon_baseline_2020,chen_closer_2019}.

\mypar{Few-shot object detection} The FSOD approaches can be summarized as meta-learning and fine-tuning (also called {\em transfer-learning})
based ones. Most meta-learning based FSOD approaches embrace formulations similar to those used in
mapping-based meta-learning approaches for FSL, \eg \cite{xiao2020few, zhang2021meta, li2021beyond,
chen2021dual, yan2019meta, kang2019few, perez2020incremental, yin2022sylph, zhang2022kernelized, han2022few}. 
Support feature aggregation is one of the main aspects where meta-learning-based methods differ from each other. Xiao and Marlet~\cite{xiao2020few} use both the differences
and the channel-wise multiplication of the features in addition to the combination of the features directly
for support-query aggregation.
Fan \etal~\cite{fan2020few} use attention blocks to make support and query features more
distinguishable for base and novel object classes. Zhang \etal\cite{zhang2021meta} use inter-class
correlations to highlight important support features. Li \etal\cite{li2021beyond} propose to use specialized
support and query features for classification and localization.

Recent efforts towards improving meta-learning based FSOD include complimentary techniques, mainly
to improve loss functions, feature matching, and novel class sample usage efficiency. \cite{li2021beyond} uses class
margin loss, \cite{hsieh2019one} uses margin-based ranking loss, \cite{zhang2021accurate} uses
hybrid loss which consist of focal loss, adaptive margin loss and contrastive loss.
Hu \etal\cite{hu2021dense} perform feature matching between query and support images to use the information
from the support images more effectively.  Similarly, Han \etal\cite{han2021query} construct a matching
network between query and support instances using heterogeneous graph convolutional networks. Li and Li~\cite{li2021transformation} augment novel class samples via adding Gaussian noise. Yin \etal~\cite{yin2022sylph} decouple classification task from localization by using the proposed class-conditional architecture.

Fine-tuning-based methods typically freeze parts of a pre-trained detection network, add auxiliary detection heads, increase the novel class variances and then apply gradient descent based model update steps, unlike meta-learning-based methods that use complex episodic learning~\cite{wang2020frustratingly, sun2021fsce, wu2020multi, fan2021generalized, kaul2022label, qiao2021defrcn, wu2021generalized}.
Wang \etal~\cite{wang2020frustratingly} propose a Faster-RCNN~\cite{ren2015faster} based approach,
where the class-agnostic region proposal network (RPN) component is kept frozen during fine-tuning.
Sun \etal\cite{sun2021fsce} use a similar approach and differently include
FPN and RPN layers to the learnable parameter set in the same architecture. These learnable layers allow using contrastive proposal encodings that facilitate the more accurate classification of novel objects. Wu \etal\cite{wu2020multi} show that the scale distribution of
support set tends to be imbalanced, and proposes a multi-scale positive sample refinement (MPSR)
branch as an addition to the main model. Fan \etal\cite{fan2021generalized} propose Retentive R-CNN
architecture to prevent forgetting during fine-tuning for base classes. The obtained
object proposals are fed into two ROI detectors responsible for base class and novel class
instances. Qiao \etal~\cite{qiao2021defrcn} focus on decoupling network modules, and introduce a gradient decoupling layer and prototypical calibration block. Kaul \etal~\cite{kaul2022label} extend the novel class annotations in the training set. In this context, the proposed method obtains object candidates from the base detector, and applies the box refinement step.

While our approach is based on fine-tuning based FSOD, we embrace meta-learn principles to optimize the loss function and augmentations to improve
the fine-tuning process for FSOD, without learning a complex and over-fitting-prone meta-model. 
The resulting loss function and data augmentations are then utilized within the fine-tuning steps.

\mypar{Automated loss function discovery} Loss function discovery is an emerging AutoML topic towards 
improving the learning systems in a data-driven manner.
Existing methods are mainly based on either
(i) constructing the loss function directly from the basic 
operators~\cite{liu2021loss, real2020automl, gonzalez2020improved} or (ii) optimizing 
parameterized loss functions~\cite{li2019lfs, wang2020loss}. For loss construction, \cite{liu2021loss}
proposes a genetic algorithm that consists of loss function verification and quality filtering
modules. In this approach, the predefined proxy task eliminates divergent and poor
candidate loss functions and survives the promising loss functions for other
steps. \cite{gonzalez2020improved} uses a genetic algorithm to select candidate loss functions from
a tree of simple mathematical operations, and the successful loss functions pass to
other stages to mutate. \cite{real2020automl} suggests a method to learn not only the loss function
but also the whole machine learning algorithm from scratch.
For loss optimization,
\cite{li2019lfs} re-analyzes the existing loss functions and presents them in a combined
formula. 
\cite{wang2020loss} observes  that the search space used in \cite{li2019lfs} can be too complex, and
propose to simplify the search space via heuristics.
In contrast to these works
targeting supervised training scenarios, we aim to adapt loss function
learning principles to the FSOD problem.

\mypar{AutoML for data augmentation} A variety of automated data augmentation techniques have recently been proposed~\cite{ho2019population, cubuk2018autoaugment, lim2019fast, cubuk2020randaugment}. Cubuk \etal~\cite{cubuk2018autoaugment} generate augmentation policies using reinforcement learning and a controller RNN.
Ho \etal~\cite{ho2019population} propose a method that reduces the computational costs compared to \cite{cubuk2018autoaugment} by using a population-based framework. Similarly, Lim \etal~\cite{lim2019fast} propose a direct Bayesian method 
to reduce costs. Cubuk \etal~\cite{cubuk2020randaugment} show that the optimal augmentation magnitudes tend to be similar across transformations, and 
the search process can greatly be simplified by using a shared value. 
We follow this suggestion and use a shared magnitude across the transforms in our formulation. 
In contrast to these works on supervised learning, however, we focus on learning detectors with few-samples.

In summary, while loss function and augmentation discovery topics increasingly attract attention
towards improving supervised training pipelines, ours is the first work on learning few-sample
specific inductive biases for fine-tuning based few-shot object detection based on meta-learning and
AutoML principles, to the best of our knowledge.

%% file: method.tex
\section{Method}

This section provides a brief summary of the FSOD problem definition and the baseline model we utilize.
We then present our definition and instantiation of 
meta-tuning.

\mypar{Problem definition} We follow the FSOD setup of~\cite{kang2019few}, where a relatively large set of training
images for the set $C_b$ of {\em base} classes is made available.
Each training image corresponds to a tuple $(x,y)$ consisting of image $x$ and annotations $y=\{ y_0, ..., y_M \}$. Each object annotation $y_i = \{c_i, b_i\}$ contains a category label ($c_i$) and a bounding box $(b_i=\{x_i, y_i, w_i, h_i\})$. Once the FSOD model training is complete,
the evaluation is carried out based on a limited number ($k$) of training images made available for the set $C_n$ of distinct {\em novel} (\ie few-shot) classes.

\mypar{Base model} We use the MPSR FSOD method~\cite{wu2020multi} as the infrastructure for our loss function and data augmentation search methods.
MPSR adapts the Faster-RCNN to be suitable for fine-tuning-based FSOD and uses an auxiliary
multi-scale positive sample refinement (MPSR) branch to handle the scale scarcity problems. This
branch expands the scale space of positive samples without increasing improper negative instances,
unlike feature pyramid networks and image pyramids that do not change data distribution, hence
the scale sparsity problem. In this context, objects in the images are cropped and resized in
multiple sizes to create scale pyramids. The MPSR uses two groups of loss functions for the region
proposal network (RPN) and detection heads, and feeds differently scaled positive samples to these
loss functions together with the main detection branch. 
Finally, we note that the proposed approach can in principle be applied to virtually any fine-tuning
based FSOD model.

\subsection{Meta-tuning loss functions}
\label{method:loss_function}

Our main goal is to improve few-shot detector fine-tuning
based on meta-learning principles. For meta-tuning the FSOD loss, we specifically focus on the 
classification loss term, as the FSOD errors tend to be primarily caused by
misclassifications~\cite{sun2021fsce}. The MPSR classification loss term can be expressed as follows:
\begin{equation} 
    \ell_{cls}(x,y) = -\frac{1}{N_{ROI}}\sum_{i}^{N_{ROI}} \log\left( \frac{ e^{f(x_i,y_i)} }{ \sum_{y} e^{f(x_i,y)} } \right)
\end{equation}
where $N_{ROI}$ is the number of ROIs (\ie candidate regions) in an image, $y_i$ is the 
groundtruth class label for the $i$-th ROI, and $f(x_i,y)$ is the corresponding class $y$ prediction score.
To add more flexibility into the loss function, we re-define it as a parametric function $\ell_{cls}(x,y;\rho)$, where
$\rho$ represents the loss function parameters. First, we introduce a temperature scalar $\rho_\tau$, \ie $\rho=(\rho_\tau)$:
\begin{equation} 
    \ell_{cls}(x,y;\rho) = -\frac{1}{N_{ROI}}\sum_{i}^{N_{ROI}} \log\left( \frac{ e^{f(x_i,y_i)/\rho_\tau} }{ \sum_{y^\prime} e^{f(x_i,y^\prime)/\rho_\tau} } \right)
\end{equation}
Our motivation 
comes from the observations on the importance of
temperature scaling in log loss on various other problems, such as knowledge
distillation~\cite{hinton2015distilling}, few-shot
classification~\cite{tadam,Ye2020FewShotLV}, and zero-shot learning~\cite{liu_generalized_2018}. While temperature
is typically tuned in a manual manner, here we aim to meta-learn it specifically for fine-tuning based FSOD purposes, giving a chance to observe the behavior of meta-tuning in a simple case. 
We also define a more sophisticated variant of the loss function by defining the {\em dynamic temperature} function $f_\rho$ and {\em novel class scaling} $\alpha$:
\begin{equation} 
    \ell_{cls}(x,y;\rho) = \frac{-1}{N_{ROI}}\sum_{i}^{N_{ROI}} \log\left( \frac{ e^{ \alpha(y_i) f(x_i,y_i)/f_\rho(t)} }{ \sum_{y^\prime} e^{ \alpha(y^\prime) f(x_i,y^\prime)/f_\rho(t)} } \right)
\end{equation}
where
%
%
    $f_\rho(t) = \exp( \rho_a t^2 + \rho_b t + \rho_c )$.
%
%
Here, $\rho=(\rho_a,\rho_b,\rho_c)$ is a 3-tuple of polynomial coefficients, and $t \in [0,1]$ is the normalized fine-tuning iteration index.
The temperature can increase or decrease over time, making the predicted class distributions smoother or sharper.
$\alpha(y)$ is set to $1$ for $y \in C_b$, and otherwise the novel class score scaling coefficient $\rho_\alpha$, as a way to learn
base and novel score balancing.

\input{fig_method}

\subsection{Meta-tuning augmentations}
\label{method:augmentation_magn}

For meta-tuning augmentations, we focus on the photometric augmentations that
are likely to be transferable from base to novel classes.
In this context, we model the brightness, saturation, contrast, and hue transforms,
with a shared magnitude parameter ($\rho_{aug}$), which is known to be effective for supervised training \cite{cubuk2020randaugment}.

\subsection{Meta-tuning procedure}

In our work, we utilize a
REINFORCE~\cite{williams1992simple} based reinforcement learning (RL) approach to search for the
optimal loss function and augmentations, where we use the AutoML approach of Wang \etal~\cite{wang2020loss}
on loss function search for fully-supervised face recognition as our starting point.

In order to meta-tune the loss function and augmentations to maximize FSOD generalization abilities, we
generate {\em proxy tasks} over base class training data to imitate real FSOD tasks over the novel
classes. For this purpose, we divide base classes into two subsets, proxy-base $\Cpxb$ and
proxy-novel $\Cpxn$. We then construct three non-overlapping data set splits using the base class training set:
(i) $\Dpxpretrain$ containing $\Cpxb$-only samples, used for training a temporary object detection model for meta-tuning purposes; (ii) $\Dpxsupport$ containing 
samples of $\Cpxb \cup \Cpxn$ classes to be used as fine-tuning images during meta-tuning; (iii)
$\Dpxquery$ containing samples of $\Cpxb \cup \Cpxn$ classes to be used for evaluating the
generalized FSOD performance of a fine-tuned model during meta-tuning.

We generate a series of FSOD proxy tasks for meta-tuning, similar to episodic meta-learning: at each proxy task $T$,
we sample a few-shot training set 
from $\Dpxsupport$. We also sample a loss function/augmentation magnitude parameter combination $\rho$,
where each $\rho_j \in \rho$ is modeled in terms of a Gaussian distribution:
%
%
    $\rho_j \sim \mathcal{N}(\mu_j,\sigma^{2})$.
%
%
Using the loss function or augmentations corresponding to the sampled $\rho$, we fine-tune
the initial model on the support images using gradient-based optimization, and compute the mean average precision (mAP)
scores on $\Dpxquery$. We get multiple mAP scores by repeating this process multiple times over multiple proxy support samples.
Meta-tuning is then carried over by updating $\mu$ values via the REINFORCE rule after each episode,
towards finding
$\mu$ values centered around well-performing $\rho$ combinations. 
\begin{equation} \label{eq5}
    \mu_j^\prime \leftarrow \mu_j + \eta R(\rho) \nabla_{\mu} \log\left(p(\rho_j; \mu_j, \sigma)\right)
\end{equation}
where $p(\rho; \mu, \sigma)$ is the Gaussian probability density function, $\eta$ is the RL learning rate.

We apply the REINFORCE update rule using the $\rho$ with the highest reward per episode.
$R(\rho)$ is the {\em normalized} reward function obtained by whitening the mAP scores.
We empirically observe that normalization improves the results (Section~\ref{sec:exp}) since without reward normalization, the 
RL updates are scaled with respect to the inherent difficulty of the proxy task, which greatly varies depending on the sampled support
examples. Reward normalization approximately removes the {\em average} reward, enabling better performing $\rho$ samples to influence  
based on their {\em relative} success.

Finally, similar to \cite{pmlr-v108-papini20a}, starting with $\sigma=0.1$, we diminish $\sigma$ over the RL iterations 
to progressively reduce explorations by sampling more conservatively, which improves converge.
The final scheme 
is illustrated in Figure~\ref{fig:method}.

%% file: fig_method.tex
\begin{figure*}[t]
\vskip 0.2in
\centering
\centerline{\includegraphics[width=\linewidth]{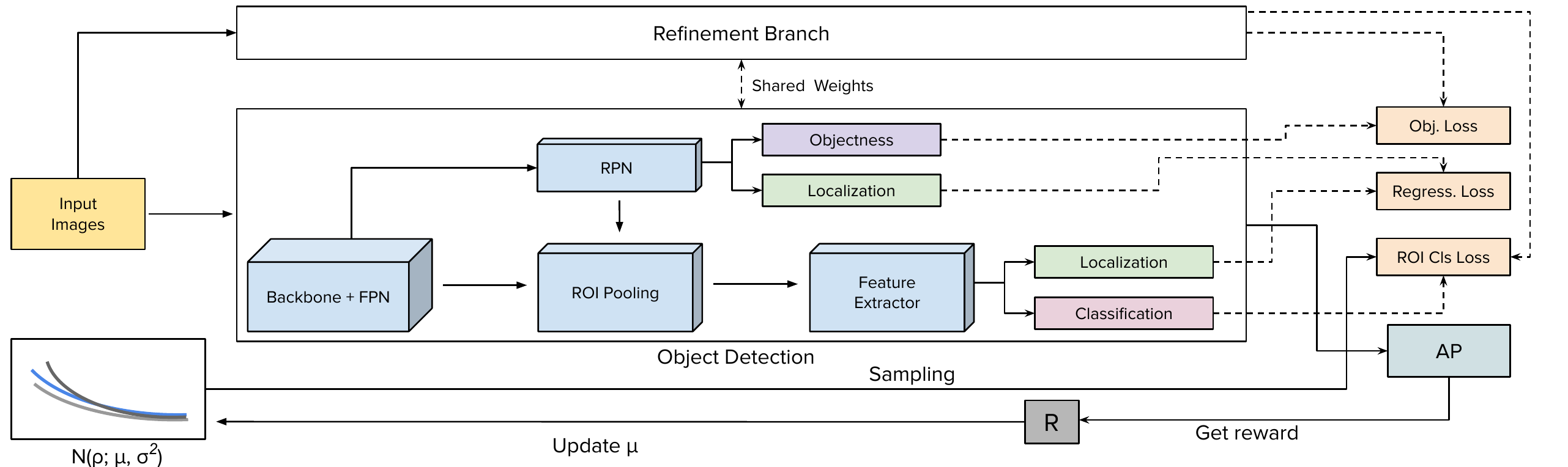}}
    \caption{The meta-tuning approach. At each RL iteration over a proxy task, the distribution parameters modeling the loss function and augmentations are updated as a function of the obtained mAP scores, towards improved training with few-samples. \label{fig:method}}
\vskip -0.2in
\end{figure*}

%% file: experiments.tex
\section{Experiments}
\label{sec:exp}

\mypar{Metrics} We use mAP to evaluate the base and novel class detection results separately. To
evaluate the generalized FSOD performance, we use the Harmonic Mean (HM) metric to compute a
balanced aggregation of base and novel class performance scores.  Adapted from generalized zero-shot
learning~\cite{xian2017zero}, HM is defined as the harmonic mean of $\text{mAP}_\text{base}$ and
$\text{mAP}_\text{novel}$ scores.

\input{avg_mpsr.tex}

\mypar{Datasets} We use Pascal VOC~\cite{everingham2010pascal} and MS
COCO~\cite{lin2014microsoft} with the same splits defined in FSOD
benchmarks~\cite{wang2020frustratingly, wu2020multi}.
On Pascal VOC, three separate base/novel class splits exist, where each one consists
of $15$ base and $5$ novel classes. In each split, we select 5 base classes to mimic novel classes during meta-tuning.  
On MS-COCO, we select 15 base classes to mimic novel classes in each proxy task,
and evaluate the models for the $10$-shot and $30$-shot settings. 

\mypar{Baselines} We primarily use the MPSR~\cite{wu2020multi} and DeFRCN~\cite{qiao2021defrcn} as our
baselines, which are among the best performing fine-tuning based FSOD methods on Pascal VOC.
For the DeFRCN experiments, we transfer the meta-tuned loss functions and augmentation magnitudes from MPSR to the DeFRCN method, which are both based on Faster-RCNN.
We take the results for FRCN~\cite{yan2019meta}, 
Ret. R-CNN~\cite{fan2021generalized}, Meta-RCNN~\cite{yan2019meta}, FSRW~\cite{kang2019few},
MetaDet~\cite{wang2019meta}, FsDetView~\cite{xiao2020few} and ONCE~\cite{perez2020incremental}
from \cite{fan2021generalized} for a fair comparison. For the
MPSR, DeFRCN (\textit{seed} is set to $0$)
and FSCE~\cite{sun2021fsce}, we report the results we obtain experimentally.  We take the results for
TFA+Hal~\cite{zhang2021hallucination}, CME~\cite{li2021beyond}, TIP~\cite{li2021transformation},
DCNet~\cite{hu2021dense}, QA-FewDet~\cite{han2021query} FADI~\cite{cao2021few}, LVC~\cite{kaul2022label}, KFSOD~\cite{zhang2022kernelized} and FCT~\cite{han2022few} from the 
original papers. 
Finally, while it is difficult to fairly compare fine-tuning versus meta-learning based approaches,
we provide a discussion in the supplementary material.

\mypar{Implementation details} We use 200 RL episodes for loss function meta-tuning, with REINFORCE
learning rate set to $0.0005$. The meta-tuning for augmentation parameter is carried out using the trained and frozen the loss function parameters. We keep the fine-tuning implementation details of MPSR unchanged, which uses 
$4000$ and $8000$ gradient descent iterations for $10$-shot and $30$-shot experiments on MS-COCO, and 
$2000$ iterations on Pascal VOC.
We will publish the full
source code upon publication; a preliminary version is provided as supplementary material.

\subsection{Main results}

We first compare the meta-tuning results against the corresponding MPSR baseline in Table~\ref{mpsr_baseline}. 
In the table, {\em Meta-Static}, {\em Meta-Dynamic}, {\em Meta-ScaledDynamic} refer to meta-tuning a single temperature, dynamic temperature, and novel class scaled dynamic temperature functions, respectively.
Similarly, {\em Aug}, {\em Meta-Static+Aug}, {\em Meta-Dynamic+Aug}, and {\em Meta-ScaledDynamic+Aug} 
refer to meta-tuning  only augmentation, single temperature and augmentation, dynamic temperature and augmentation, and novel class scaled dynamic temperature and augmentation functions, respectively.
We observe that meta-tuning consistently improves the FSOD and G-FSOD results of the MPSR model. We also observe steady improvements gradually from the baseline to Meta-Static, to Meta-Dynamic, and finally to Meta-
ScaledDynamic. In addition, the meta-tuned augmentation magnitude parameter also contributes positively to the few-shot object detection performance. The overall consistency of the improvements provides positive evidence for the value of loss and augmentation meta-tuning. 

\mypar{Pascal VOC results} In Table~\ref{pascal_avg_tl}, 
we report the Pascal VOC results for our MPSR and DeFRCN based Meta-ScaledDynamic+Aug approach and compare them against the state-of-the-art fine-tuning
based FSOD methods. While we present the scores averaged over the three splits in this table,
additional per-split FSOD and G-FSOD results can be found in the supplementary material.
The left side of Table~\ref{pascal_avg_tl} presents the FSOD results for the varying number of support images.
We observe that DeFRCN combined with Meta-ScaledDynamic+Aug, \ie meta-tuning of the score coefficient, dynamic temperature and the augmentation parameter, yields the best mAP scores in all $k$-shot settings among all methods.

\input{pascal_avg_tl}

The right side of Table~\ref{pascal_avg_tl} presents the G-FSOD results on Pascal VOC.
We observe that the best-performing Meta-ScaledDynamic+Aug method improves the HM scores further above the state-of-the-art in all $k$-shot settings. 
Overall, these results 
suggest that the proposed framework
is an effective way for meta-learning inductive biases to be used in fine-tuning-based FSOD.

Figure~\ref{fig:visual_comparison} presents visual detection examples without and 
with meta-tuned scaled dynamic temperature and augmentations in the first and second rows, respectively. We observe various improvements, such as reductions in false positives, improved recall, and more precise boxes, most likely due to the improved model fitting in the low-data regime.

\input{visual-comparison}

\mypar{MS-COCO results} In Table~\ref{coco_novel_tl}, we compare the MPSR and DeFRCN based Meta-ScaleDynamic+Aug results against other fine-tuning based FSOD methods that report
$10$-shot and $30$-shot results on the MS-COCO dataset. We observe that with meta-tuning, the FSOD scores of MPSR improve from $9.1$ to
$12.5$ ($10$-shot mAP), and from $13.7$ to $15.4$ ($30$-shot mAP). We also observe that the scores of DeFRCN improve from $18.5$ to $18.8$ ($10$-shot mAP), and from $21.9$ to $23.4$ ($30$-shot mAP), obtaining the best and second best results against all other models. Similarly, in the case of G-FSOD, with
meta-tuning, the $10$-shot HM score of DeFRCN improves from $24.0$ to $24.4$, outperforming all other models. In addition, the $30$-shot HM score of DeFRCN improves from $26.8$ to $28.0$, which is slightly below the $28.1$ score of LVC-PL~\cite{kaul2022label}.

\input{coco_novel_tl}

\input{proxy}

\subsection{Ablation studies}
\label{ablation}

\mypar{Meta-tuning details} 
The proposed meta-tuning approach involves three important technical details: {\em Proxy-novel imitation},
{\em model re-initialization}, and {\em reward normalization}.
Proxy-novel imitation refers to reinforcement learning over the sampled proxy-novel tasks, instead of 
the whole training set, to mimic the test-time FSOD challenges.
Model re-initialization is the re-initialization of the base model for each task.
Without re-initialization, not only the sampled loss/augmentation parameters and tasks but also the accumulated
model updates undesirably affect the rewards. Reward normalization further reduces the effect of 
task difficulty variance by normalizing the rewards obtained within a single episode, allowing
a more isolated assessment of the sampled loss functions and augmentations.

\input{time-mu}

We evaluate the contributions of these three important details in terms of G-FSOD
HM scores using the $5$-shot setting of Pascal VOC Split-1 with 
MPSR+Meta-Dynamic. The results averaged over $5$ runs are given in Table~\ref{proxy_task}.
We observe that each component progressively improves the HM scores, and the most significant contribution is
made by reward normalization, which improves from $62.1$ to $63.3$.
We also observe that reward normalization considerably improves the overall experimental stability.
To quantify this observation, we estimate the 95$\%$ confidence interval over the runs using $CI =
1.96\frac{s}{\sqrt{n}}$, where  $s$, $n$, and $1.96$ are the standard deviation, number of
runs, and $Z$-value, respectively~\cite{wang2020frustratingly}. According to this estimator,
the normalization step narrows the confidence interval from $\pm 0.75$ to $\pm 0.13$, providing a 
clear improvement in reliability.

\mypar{Learned loss functions} In Figure~\ref{fig:time-mu}, we plot the learned loss functions 
according to the $\mu$ values obtained at the end of the RL process. The upper plot  
shows the dynamic temperature functions learned over three different splits.
We observe that temporally attenuated temperature values are preferred consistently, sharpening the
predictions towards the end of the fine-tuning process. The lower plot shows the
learned dynamic temperature functions with novel class score scaling. The learned scaling
coefficients, \ie $\mu_\alpha$ of the learned $\rho_\alpha$ distribution, are shown as horizontal lines. We observe that similar 
dynamic temperature functions are learned, and $\mu_\alpha$ values vary between $1.09$ to $1.2$, suggesting that
the meta-tuning process learns to boost the novel class scores. The interpretability of these outcomes, we believe, highlights a significant advantage of loss meta-tuning. In the context of interpretability, we observe that as the fine-tuning process continues on the few-shot training set, the predictions are progressively made sharper, \textit{i.e.} the loss becomes more sensitive to classification errors and enforces towards making more confident correct predictions. This is in alignment with one of our original motivations for reducing the dominating classification errors in G-FSOD, as the meta-tuning process automatically learns to enforce more accurate classifications, where the curve steepness and the numerical ranges are learned via RL.

\input{low_shot_coco} 

\mypar{Learned augmentations} The learned photometric augmentation magnitude values learned are $0.29$, $0.24$, $0.13$, and $0.36$ for Pascal VOC split-1, split-2, split-3, and MS-COCO datasets, respectively. We observe that the learned augmentation magnitudes positively contribute to the performance. According to the results in Table~\ref{mpsr_baseline}, the average  Pascal VOC split-1/1-shot score increases from $33.1$ to $34.6$ with only augmentation steps.

\mypar{Very low-shot experiments} Finally, we evaluate the meta-tuning approach in low-shot many-class settings.~\cite{zhang2021hallucination} proposes
TFA+Hal method that uses the TFA baseline and conducts $1$-shot, $2$-shot, and $3$-shot FSOD on the MS-COCO
dataset. As we already observe the positive effects of the loss terms and augmentation magnitudes obtained from the MPSR on the DeFRCN, we similarly apply the learned parameters to the TFA baseline. The results are presented in Table~\ref{low_shot_coco}. We observe that results are consistently  improved using the meta-tuned functions on the TFA baseline.

%% file: avg_mpsr.tex
\begin{table*}[]
\begin{scriptsize}
\centering
\begin{tabular}{c|cccccccccc|cccc}
\toprule
\multirow{3}{*}{Method/Shot} & \multicolumn{10}{c|}{Pascal VOC}                                                                                                                                                   & \multicolumn{4}{c}{MS-COCO}                                                               \\
                             & \multicolumn{5}{c}{Novel Classes}                                                                  & \multicolumn{5}{c|}{All Classes (HM)}                                         & \multicolumn{2}{c}{Novel Classes}                  & \multicolumn{2}{c}{All Classes (HM)} \\
                             & 1             & 2             & 3             & 5             & \multicolumn{1}{c|}{10}            & 1             & 2             & 3             & 5             & 10            & 10            & \multicolumn{1}{c|}{30}            & 10                & 30               \\ \midrule
MPSR~\cite{wu2020multi}                         & 33.1          & 37.2          & 44.3          & 47.1          & \multicolumn{1}{c|}{52.1}          & 43.1          & 47.4          & 54.5          & 57.2          & 60.8          & 9.1           & \multicolumn{1}{c|}{13.7}          & 11.5              & 15.0             \\
MPSR+Meta-Static             & 33.4          & 39.4          & 45.1          & 47.3          & \multicolumn{1}{c|}{52.6}          & 43.7          & 50.4          & 55.4          & 57.5          & 61.4          & 10.1          & \multicolumn{1}{c|}{14.8}          & 12.7              & 16.4             \\
MPSR+Meta-Dynamic            & 34.5          & 39.8          & 45.0          & 48.2          & \multicolumn{1}{c|}{52.5}          & 45.0          & 51.0          & 55.5          & 58.3          & 61.6          & 11.9          & \multicolumn{1}{c|}{14.9}          & 14.3              & 16.6             \\
MPSR+Meta-ScaledDynamic      & 35.2          & 40.3          & 45.8          & 48.4          & \multicolumn{1}{c|}{52.9}          & 45.6          & 51.2          & 55.9          & 58.3          & 61.8          & 12.3          & \multicolumn{1}{c|}{15.0}          & 14.4              & 16.7             \\ \midrule
MPSR+Aug                       & 34.6          & 38.6          & 46.0          & 48.3          & \multicolumn{1}{c|}{52.7}          & 45.1          & 49.5          & 56.2          & 58.4          & 62.0          & 9.9           & \multicolumn{1}{c|}{14.9}          & 12.5              & 16.3             \\
MPSR+Meta-Static+Aug           & 35.3          & 39.1          & 46.1          & 48.4          & \multicolumn{1}{c|}{52.7}          & 45.9          & 49.9          & 56.2          & 58.3          & 61.8          & 10.2          & \multicolumn{1}{c|}{15.2}          & 12.8              & 16.7             \\
MPSR+Meta-Dynamic+Aug          & 35.4          & 39.6          & 46.5          & 48.9          & \multicolumn{1}{c|}{53.3}          & 46.0          & 50.5          & 56.8          & 58.9          & 62.5          & 12.1          & \multicolumn{1}{c|}{15.3}          & 14.5              & 16.8             \\
MPSR+Meta-ScaledDynamic+Aug    & \textbf{35.8} & \textbf{40.6} & \textbf{46.8} & \textbf{49.2} & \multicolumn{1}{c|}{\textbf{53.7}} & \textbf{46.3} & \textbf{51.5} & \textbf{57.0} & \textbf{59.2} & \textbf{62.7} & \textbf{12.5} & \multicolumn{1}{c|}{\textbf{15.4}} & \textbf{14.7}     & \textbf{16.9}   \\
\bottomrule
\end{tabular}
\caption{FSOD (mAP) and G-FSOD (HM of the base and novel class mAPs) results on Pascal VOC and MS-COCO datasets for MPSR baseline method. HM stands for harmonic mean.}
\label{mpsr_baseline}
\end{scriptsize}
\end{table*}

%% file: pascal_avg_tl.tex
\begin{table*}[t]
\centering
\begin{scriptsize}
\begin{tabular}{c|ccccc|ccccc}
\toprule
\multirow{2}{*}{Method/Shot} & \multicolumn{5}{c|}{Novel Classes}&  \multicolumn{5}{c}{All Classes (HM)} \\
                             & 1             & 2             & 3             & 5             & 10             & 1             & 2             & 3             & 5             & 10            \\ \midrule
FRCN~\cite{yan2019meta}~~~(ICCV'19)       & 16.1                  & 20.6                  & 28.8                  & 33.4                  & 36.5 & 25.9          & 31.7          & 40.0          & 44.3          & 46.7                  \\
                      TFA-fc~\cite{wang2020frustratingly}~~~(ICML'20)     & 27.6                  & 30.6                  & 39.8                  & 46.6                  & 48.7   & 40.5          & 44.1          & 52.9          & 58.3          & 59.9                \\
                      TFA-cos~\cite{wang2020frustratingly}~~~(ICML'20)     & 31.4                  & 32.6                  & 40.5                  & 46.8                  & 48.3   & 44.6          & 46.0          & 53.5          & 58.4          & 59.6                \\
                      FSCE~\cite{sun2021fsce}~~~(CVPR'21)      & 29.2                  & 36.3                  & 42.5                  & 47.1                 & 52.2 & 41.8 & 48.8          & 54.2          & 57.7 & 61.0
                      
                        \\
                      Ret. R-CNN~\cite{fan2021generalized}~~~(CVPR'21) & 31.4                  & 37.1                  & 41.4                  & 46.8                  & 48.8         & 44.7 & 50.5          & 54.7          & 59.1 & 60.8          \\
                      TFA+Hal~\cite{zhang2021hallucination}~~~(CVPR'21)     & 32.9                  & 35.5                  & 40.4                  & 46.3                  & 48.1     & - & - & - & - & -            \\
                      FADI~\cite{cao2021few}~~~(NeurIPS'21)     & 42.2                 & 46.5                  & 47.9                  & 52.4                 & 56.9     & - & - & - & - & -            \\
                                            LVC~\cite{kaul2022label}~~~(CVPR'22)       & 30.9& 35.4	& 43.6	& 51.1	& 54.1&-&-&-&-&-            \\
                      LVC-PL ~\cite{kaul2022label}~~~(CVPR'22)      & 45.2&	45.0	&54.8&	57.5	&58.6&-&-&-&-&-            \\
                      MPSR~\cite{wu2020multi}~~~(ECCV'20)       & 33.1                  & 37.2                  & 44.3                  & 47.1                  & 52.1        & 43.1          & 47.4          & 54.5          & 57.2          & 60.8            \\
                      DeFRCN~\cite{qiao2021defrcn}~~~(ICCV'21)       & \textcolor{blue}{\textbf{46.5}}	& \textcolor{blue}{\textbf{52.6}}	& \textcolor{blue}{\textbf{55.9}}& \textcolor{blue}{\textbf{60.0}} & \textcolor{blue}{\textbf{60.8}} & \textcolor{blue}{\textbf{57.6}}	& \textcolor{blue}{\textbf{62.5}}	& \textcolor{blue}{\textbf{64.7}}	& \textcolor{blue}{\textbf{67.6}} & \textcolor{blue}{\textbf{67.8}}            \\
                       \midrule
                      MPSR+Meta-ScaledDynamic+Aug  & 35.8	& 40.6	& 46.8	& 49.2	&53.7 & 46.3	& 51.5	& 57.0	& 59.2	& 62.7\\ 
                      DeFRCN+Meta-ScaledDynamic+Aug  & \textcolor{red}{\textbf{49.2}}                  & \textcolor{red}{\textbf{54.0}}        & \textcolor{red}{\textbf{57.2}}                  & \textcolor{red}{\textbf{61.3}}         & \textcolor{red}{\textbf{61.8}} & \textcolor{red}{\textbf{59.8}}          & \textcolor{red}{\textbf{63.7}} & \textcolor{red}{\textbf{65.9}} & \textcolor{red}{\textbf{68.6}}         & \textcolor{red}{\textbf{68.7}}\\
\bottomrule
\end{tabular}
\caption{FSOD (mAP) and G-FSOD (HM of the base and novel class mAPs) results on Pascal VOC. The best and the second-best results are marked with \textcolor{red}{red} and \textcolor{blue}{blue}. HM stands for harmonic mean.}
\label{pascal_avg_tl}
\end{scriptsize}
\end{table*}

%% file: visual-comparison.tex
\begin{figure*}
\centering
\resizebox{\textwidth}{!}{%
\begin{tabular}{cccccc}
\HangBox{\includegraphics[width=3.8cm, height=3.4cm]{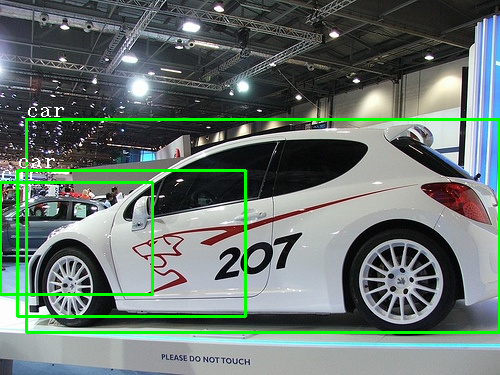}}
\HangBox{\includegraphics[width=3.8cm, height=3.4cm]{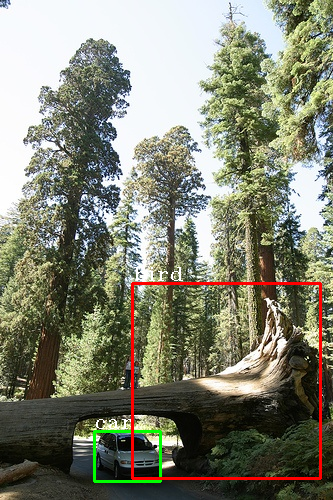}}
\HangBox{\includegraphics[width=3.8cm, height=3.4cm]{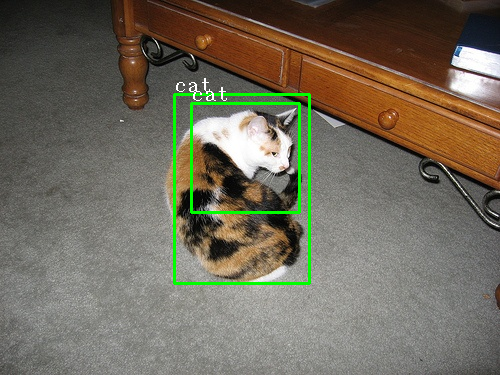}}
\HangBox{\includegraphics[width=3.8cm, height=3.4cm]{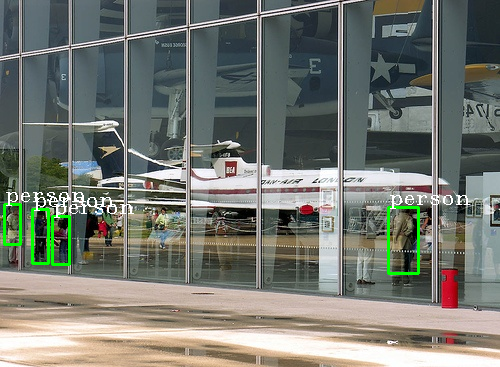}}
\HangBox{\includegraphics[width=3.8cm, height=3.4cm]{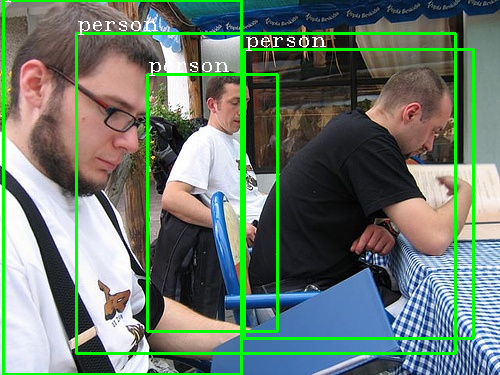}}
\HangBox{\includegraphics[width=3.8cm, height=3.4cm]{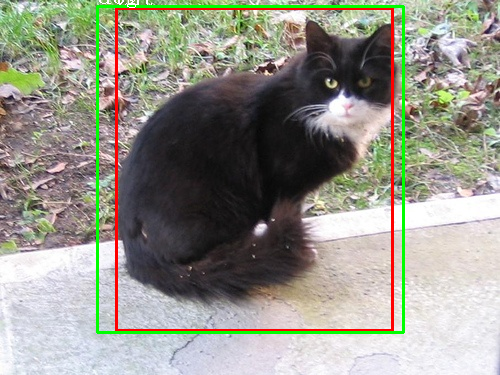}}
\\
\HangBox{\includegraphics[width=3.8cm, height=3.4cm]{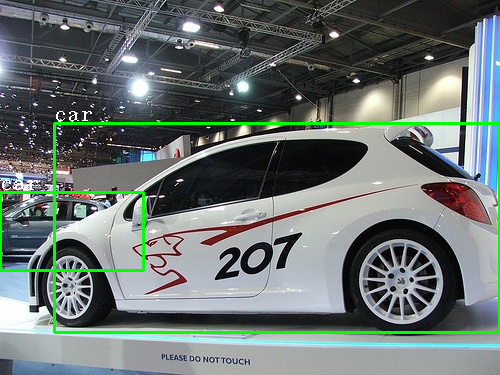}}
\HangBox{\includegraphics[width=3.8cm, height=3.4cm]{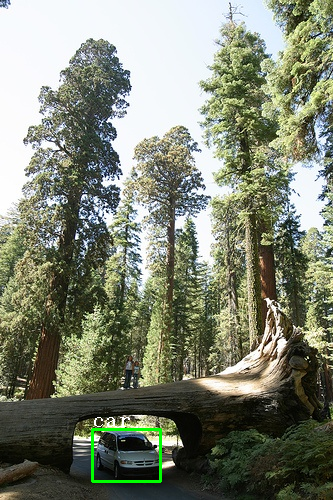}}
\HangBox{\includegraphics[width=3.8cm, height=3.4cm]{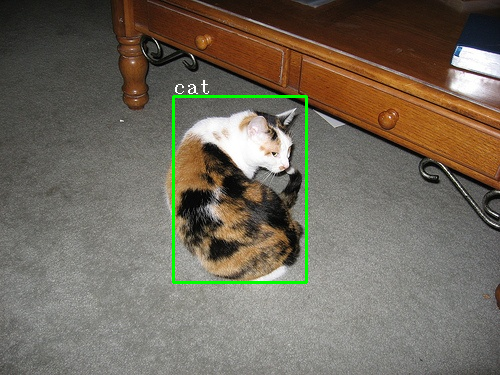}}
\HangBox{\includegraphics[width=3.8cm, height=3.4cm]{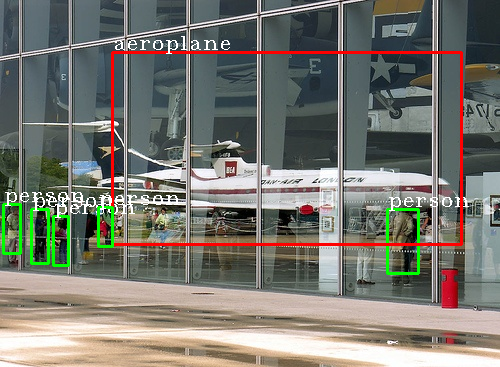}}
\HangBox{\includegraphics[width=3.8cm, height=3.4cm]{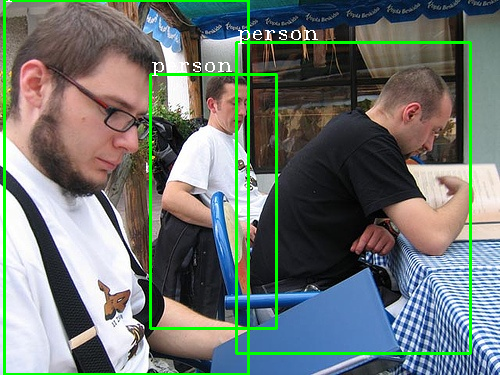}}
\HangBox{\includegraphics[width=3.8cm, height=3.4cm]{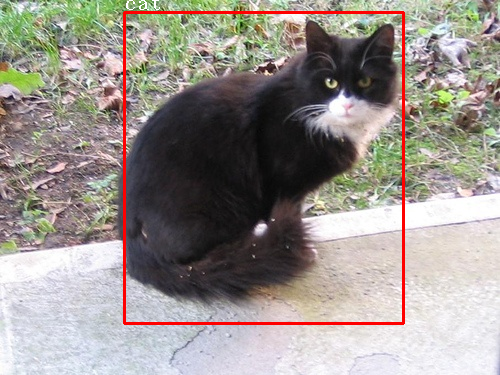}}\\
Split-1/5-shot~~~~~~~~~~~~~~~~~~~~~Split-1/10-shot~~~~~~~~~~~~~~~~~~~~~Split-2/5-shot~~~~~~~~~~~~~~~~~~~~~Split-2/10-shot~~~~~~~~~~~~~~~~~~~~~Split-3/5-shot~~~~~~~~~~~~~~~~~~~~~Split-3/10-shot
\end{tabular}
}
\caption{Qualitative results using MPSR without (first row) and with (second row) meta-tuning,
over multiple Pascal VOC splits. Base and novel class detections are shown with \textcolor{green}{green} and \textcolor{red}{red} boxes, respectively. (Best viewed in color.)}
\label{fig:visual_comparison}
\end{figure*}

%% file: coco_novel_tl.tex
\begin{table}[]
\centering
\begin{scriptsize}
\begin{tabular}{c|cc|cc}
\toprule
\multirow{2}{*}{Method/Shots} & \multicolumn{2}{c|}{Novel Classes} & \multicolumn{2}{c}{All Classes (HM)} \\
                                         & 10-shot                          & 30-shot                          & 10-shot                          & 30-shot          \\ \midrule
FRCN~\cite{yan2019meta}~~~(ICCV'19)        & 9.2                              & 12.5                             & 12.8                             & 15.6                        \\
FRCN-BCE~\cite{wang2020frustratingly}~~~(ICML'20)    & 6.4                              & 10.3                             & 10.9                             & 16.1                          \\
TFA-fc~\cite{wang2020frustratingly}~~~(ICML'20)      & 10.0                             & 13.4                             & 15.4                             & 19.4                        \\
TFA-cos~\cite{wang2020frustratingly}~~~(ICML'20)     & 10.0                             & 13.7                             & 15.6                             & 19.8                           \\
MPSR~\cite{wu2020multi}~~~(ECCV'20)                  & 9.1                              & 13.7                             & 11.5                             & 15.0                         \\
    FSCE~\cite{sun2021fsce}~~~(CVPR'21)              & 10.5                             & 14.4 & 16.0                             & 20.2                           \\
    Ret. R-CNN~\cite{fan2021generalized}~~~(CVPR'21) & 10.5                             & 13.8                             & 16.6  & 20.4                                 \\
    FADI~\cite{cao2021few}~~~(NeurIPS'21) &12.2 & 16.1 & - & -\\
    DeFRCN~\cite{qiao2021defrcn}~~~(ICCV'21) & \textcolor{blue}{\textbf{18.5}} & 21.9 & \textcolor{blue}{\textbf{24.0}} & 26.8
    \\
    LVC ~\cite{kaul2022label}~~~(CVPR'22) & 12.1 & 17.8 & 17.8 & 22.8\\
    LVC-PL ~\cite{kaul2022label}~~~(CVPR'22) & 17.8 & \textcolor{red}{\textbf{24.5}} & 22.8 & \textcolor{red}{ \textbf{28.1}} \\
    
    \midrule

MPSR+Meta-ScaledDynamic+Aug                  & 12.5  & 15.4  & 14.7                             & 16.9                               \\
    DeFRCN+Meta-ScaledDynamic+Aug               & \textcolor{red}{ \textbf{18.8}}         & \textcolor{blue}{ \textbf{23.4}}  & \textcolor{red}{ \textbf{24.4}} & \textcolor{blue}{ \textbf{28.0}}
\\
\bottomrule
\end{tabular}
\caption{Comparison of Meta-ScaledDynamic results to the fine-tuning based (G-)FSOD methods on the MS-COCO dataset. The best and the second-best results are marked with \textcolor{red}{red} and \textcolor{blue}{blue}.}
\label{coco_novel_tl}
\end{scriptsize}
\end{table}

%% file: proxy.tex
\begin{table}[t]
\centering
\begin{scriptsize}
\begin{tabular}{ccc|c}
\toprule

    Proxy-novel imit. & Model re-init. & Reward norm. & HM \\
\midrule
    \xmark                                & \xmark & \xmark & 61.5\\
                                   \cmark & \xmark & \xmark & 61.8\\
                                   \cmark & \cmark & \xmark & 62.1\\
                                   \cmark & \cmark & \cmark & 63.3\\
\bottomrule
\end{tabular}
    \caption{Evaluation of meta-tuning details. \textit{Proxy-novel imitation} is the imitation of novel classes using a subset of base classes. \textit{Model re-initialization} is the re-initialization of the base model at each task. \textit{Reward normalization} is within-episode normalization of the mAP scores during meta-tuning.}
\label{proxy_task}
\end{scriptsize}

\end{table}

%% file: time-mu.tex
\begin{figure}[t!]

  \centering
  \includegraphics[width=.9\linewidth]{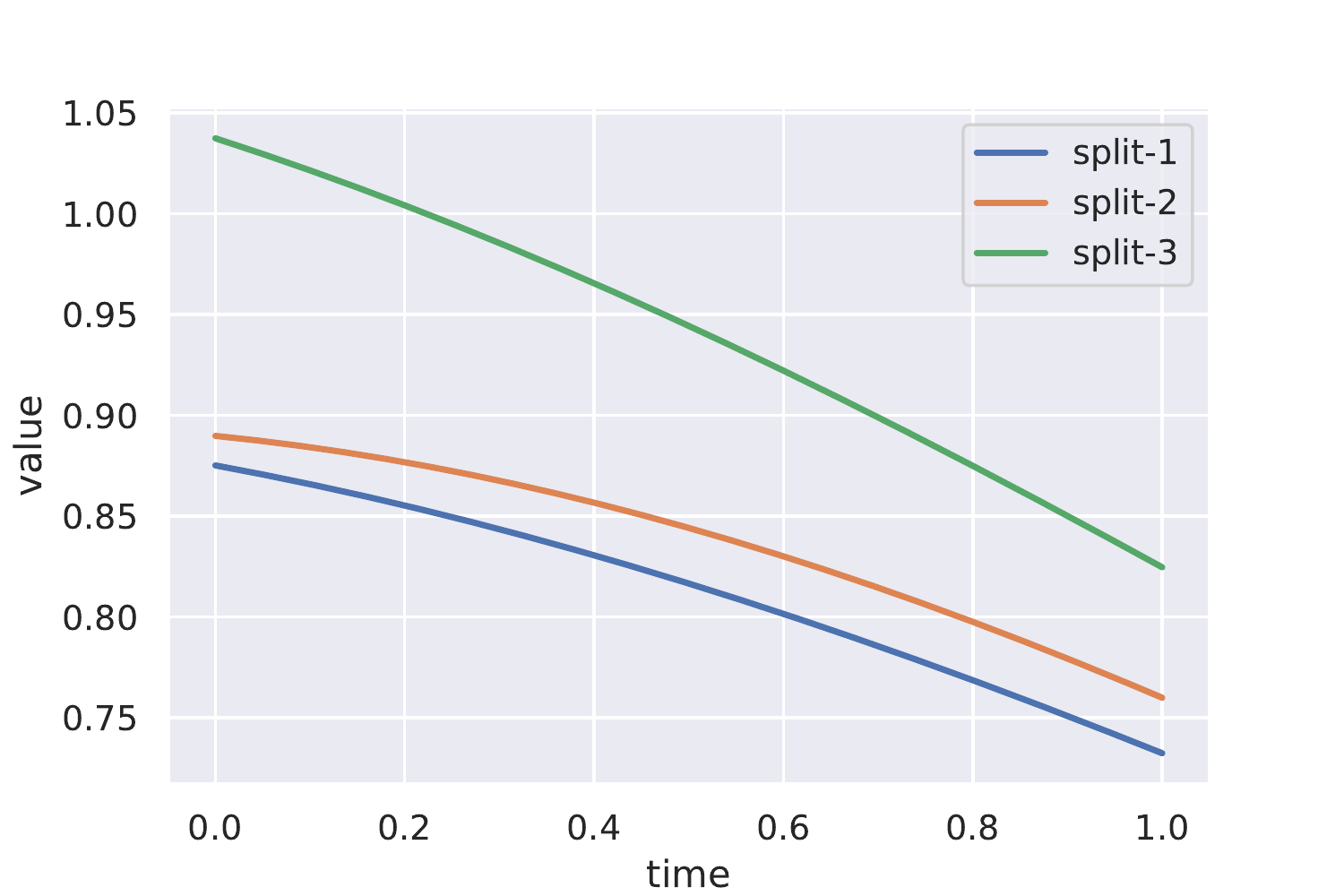}

  \includegraphics[width=.9\linewidth]{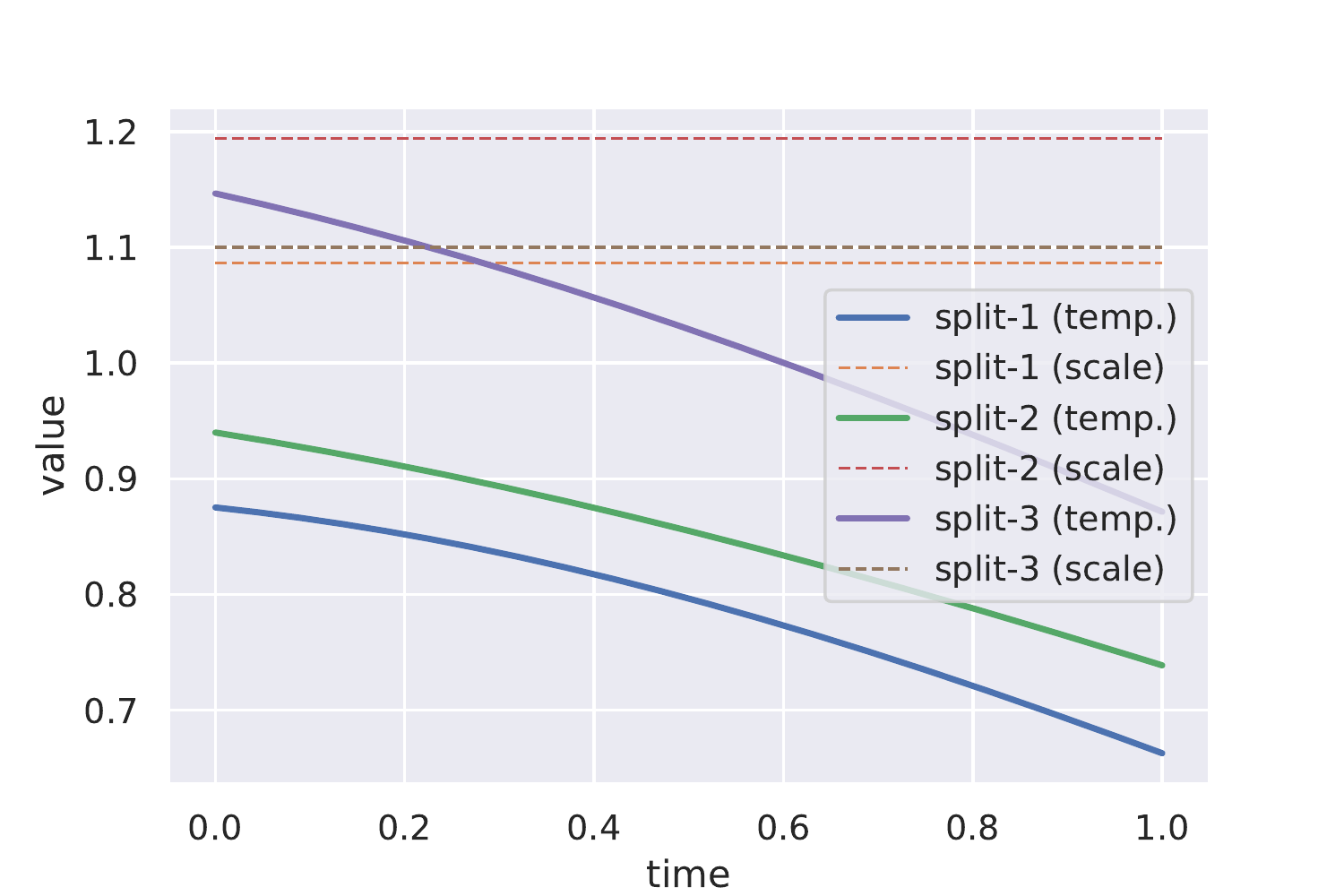}

    \caption{The dynamic temperature functions and score scaling coefficients learned by the meta-tuning process, using Meta-Dynamic (upper) and Meta-ScaledDynamic (lower) formulations. Results for each Pascal VOC split is shown with a separate curve. \label{fig:time-mu}}
\end{figure}

%% file: low_shot_coco.tex
\begin{table}
\centering
\begin{scriptsize}
\begin{tabular}{c|c|c|c}
\toprule
S/M & TFA~\cite{wang2020frustratingly} & TFA+Hal~\cite{zhang2021hallucination} & TFA+Meta-ScaledDynamic+Aug     \\ \midrule
1           & 3.4 & 3.8     & \textbf{4.7} \\
2           & 4.6 & 5.0     & \textbf{5.8} \\
3           & 6.6 & 6.9     & \textbf{7.1} \\
\bottomrule
\end{tabular}
\caption{Low-shot (1-shot, 2-shot and 3-shot) experiments on MS-COCO dataset with novel classes.}
\label{low_shot_coco}
\end{scriptsize}
\end{table}

%% file: conclusion.tex
\section{Conclusion}
\label{sec:conclusion}

Fine-tuning based frameworks offer simple and reliable approaches to building detection models from
few samples.  However, a major limitation of the existing fine-tuning-based FSOD models is their
focus on the hand-crafting the design of fine-tuning details for few-shot training, which is
inherently difficult and likely to be sub-optimal. Towards addressing this limitation, we propose to
meta-learn the fine-tuning based learning dynamics as a way of introducing learned inductive biases
for few-shot learning. The proposed tuning scheme uses meta-learning principles with reinforcement
learning, and obtains interpretable loss functions and augmentation magnitudes for few-shot
training. Our comprehensive experimental results on Pascal VOC and MS COCO datasets show that the
proposed meta-tuning approach consistently provides significant performance improvements over the strong
fine-tuning based few-shot detection baselines in both FSOD and G-FSOD settings.

While we restrict our experiments to loss and augmentation functions, meta-tuning other learning components, \eg initial model, and applications to other few-shot learning problems can be interesting future work directions.

%% file: appendix.tex
\section{Appendix}

\subsection{Proxy task class splits}
\label{app_dataset}

We use proxy tasks to apply the meta-tuning ideas, so we generate sub-splits in the base classes. In this context, we select some base classes to mimic novel classes to conduct the proxy task. We summarize the list of proxy Pascal VOC classes on Table~\ref{pascal_proxy_class_list}. The list of selected proxy novel classes for the MS-COCO dataset is as follows: \{\textit{"skis", "tennis racket", "scissors", "truck", "baseball bat", "handbag", "carrot", "mouse", "parking meter", "apple", "knife", "microwave", ""refrigerator", "cake", "zebra"}\}.

\input{pascal_proxy_classes}

\subsection{Algorithm}
\label{app_proxy}

We summarize the main meta-tuning procedure in Algorithm~\ref{app_algorithm_proxy_task_app}. We can divide this algorithm into three parts: (i) model initialization and parameter sampling, (ii) instance sampling and mAP calculation, (iii) mAP normalization and RL steps.

\mypar{1) Model initialization and parameter sampling} This algorithm firstly initializes the base proxy detection model weights for the proxy task and sample $\rho$ value from normal distributions. The base proxy detection model represents the object detection model trained using the $\Dpxpretrain$ dataset.

\mypar{2) Instance sampling and mAP calculation} 
The proposed algorithm samples new instances from the proxy fine-tuning dataset $\Dpxsupport$, and calculates the mean average precision scores on proxy validation dataset $\Dpxquery$ after a certain number of iterations. The algorithm repeats this process for N times.

\mypar{3) mAP normalization and RL steps} The proposed algorithm normalizes the mAP scores, selects the maximum score as the reward value among the normalized APs, and applies a single REINFORCE step.

\begin{algorithm}
\caption{Meta-tuning Loss Function Learning}
\label{app_algorithm_proxy_task_app}
\begin{center}
\begin{algorithmic}
   \State {\bfseries Input:} Pre-trained model $m_{init}$, proxy fine-tuning dataset $\Dpxsupport$, proxy validation dataset $\Dpxquery$, number of $rho$ trials $N$, maximum iteration number $M$
   \State
   \State iteration\_index = 1
   \Repeat
   \State Initialize $m_{init}$ and sample new $\rho$
   \For{$rho\_index=1$ {\bfseries to} $N$}
    \State Sample new fine-tuning images from $\Dpxsupport$
    \State Take $m_{init}$, run all iter. using current samples
    \State Calculate mAP[$rho\_index$] on $\Dpxquery$
 \EndFor
   \State{\bfseries end for}
   \State Normalize mAP scores 
   \State Get max normalized AP as a reward
   \State Make a single REINFORCE step
   \State iteration\_index += 1
   \Until{iteration\_index = M}
\end{algorithmic}
\end{center}
\end{algorithm}

\input{pascal_novel_tl}

\input{pascal_gfsd_tl}

\input{pascal_novel_ml}

\input{pascal_gfsd_ml}

\input{coco_novel_gfsd_ml}

\input{visual-results-appendix}

\input{visual-results-defrcn-appendix}

\subsection{Additional Experimental Results}
\label{app_exp}

In this section, we share detailed experimental comparison results for Pascal VOC and MS COCO datasets.

\mypar{Comparison to fine-tuning based FSOD and G-FSOD methods on Pascal VOC} We first present the detailed Pascal VOC comparisons for each split and shot with only novel classes in Table~\ref{pascal_novel_tl}, and the detailed comparisons with all classes in Table~\ref{pascal_gfsd_tl}. The experimental results show that the meta-tuning approach significantly improves the strong fine-tuning based few-shot detection baselines on the Pascal VOC benchmark. 
We provide complementary 
visual results of the MPSR+Meta-ScaledDynamic+Aug method using the Pascal VOC split-3/10-shot setting in Figure~\ref{fig:visual_comparison_app}. We also present examples from the 
visual results of the DeFRCN+Meta-ScaledDynamic+Aug method using the Pascal VOC split-2/10-shot setting in Figure~\ref{fig:visual_comparison_app_defrcn}.

\mypar{Comparisons to meta-learning based FSOD and G-FSOD on Pascal VOC}
We present the detailed Pascal VOC comparisons with meta-learning based methods in Table~\ref{pascal_novel_ml} and Table~\ref{pascal_gfsd_ml} for novel-only and all-classes settings, respectively. Since the most of the meta-learning methods do not share G-FSOD results, we are able to compare against a more limited number of meta-learning methods than FSOD. The experimental results (Table~\ref{pascal_novel_ml}) show that our DeFRCN+Meta-ScaledDynamic+Aug method obtains the best results in all of the FSOD cases, except for the Split-2/1-shot setting. In the G-FSOD experiments (Table~\ref{pascal_gfsd_ml}), it is observed that the proposed meta-tuning approach obtains the state-of-the-art results with a clear margin against existing meta-learning based methods.

\mypar{Comparisons to meta-learning based FSOD and G-FSOD on MS-COCO}
We compare our results with meta-learning based methods on the MS-COCO dataset and share the obtained results in Table~\ref{coco_base_novel_ml}. In this table, we are able to report a rather limited number of meta-learning methods to compare the G-FSOD results since most meta-learning based methods do not share G-FSOD results on the MS-COCO dataset. In FSOD experiments, we also observe that our DeFRCN+Meta-ScaledDynamic+Aug method obtains higher results than several recently published meta-learning based methods. We additionally observe major improvements in terms of HM scores in the G-FSOD setting, similar to the improvements obtained on the Pascal VOC dataset.

\subsection{Implementation and runtime}
\label{imp_runtime}
We run our MPSR and DeFRCN experiments on a server with 4 Nvidia Tesla V100 32GB GPUs. The base MPSR model training to be used during fine-tuning takes 0.25 days for Pascal VOC and 0.45 days for MS COCO datasets. Since the base models used for the proxy tasks contain fewer classes and demand fewer iterations, the training of the MPSR model takes 0.1 days in Pascal VOC and 0.6 days in MS COCO datasets for the proxy-base classes. RL training for meta-tuning using the final setting takes 0.05 days for Pascal VOC splits and 0.5 days for the MS COCO dataset. Finally, we note that meta-tuning operations do not incur any overhead during the fine-tuning for novel classes.

%% file: pascal_proxy_classes.tex
\begin{table*}[h]
\begin{center}
\begin{tabular}{ccc|ccc}
\toprule
    \multicolumn{3}{c|}{\textbf{Proxy-base classes ($\Cpxb$)}}             & \multicolumn{3}{c}{\textbf{Proxy-novel classes ($\Cpxn$)}}             \\
{Split-1} & {Split-2} & {Split-3} & {Split-1} & {Split-2} & {Split-3} \\ \midrule
aeroplane   & bicycle     & aeroplane   & person      & motorbike & horse       \\
bicycle     & bird        & bicycle     & pottedplant & person    & person      \\
boat        & boat        & bird        & sheep       & sheep     & pottedplant \\
bottle      & bus         & bottle      & train       & train     & train       \\
car         & car         & bus         & tvmonitor   & tvmonitor & tvmonitor   \\
cat         & cat         & car         &             &           &             \\
chair       & chair       & chair       &             &           &             \\
diningtable & diningtable & cow         &             &           &             \\
dog         & dog         & diningtable &             &           &             \\
horse       & pottedplant & dog         &             &           &     \\
\bottomrule
\end{tabular}
\caption{Proxy task class splits for Pascal VOC. }
\label{pascal_proxy_class_list}
\end{center}
\end{table*}

%% file: pascal_novel_tl.tex
\begin{table*}
\begin{center}
\begin{scriptsize}
\resizebox{\textwidth}{!}{%
\begin{tabular}{c|ccccc|ccccc|ccccc}
\toprule
\multicolumn{1}{c|}{}                      & \multicolumn{5}{c|}{Split 1}                                                                                                                                                                 & \multicolumn{5}{c|}{Split 2}                                                                                                                                                                 & \multicolumn{5}{c}{Split 3}                                                                                                                                                                  \\
\multicolumn{1}{c|}{\multirow{-2}{*}{Method/Shot}} & 1                                    & 2                                    & 3                                    & 5                                    & 10                                   & 1                                    & 2                                    & 3                                    & 5                                    & 10                                   & 1                                    & 2                                    & 3                                    & 5                                    & 10                                   \\ \midrule
                         FRCN~\cite{yan2019meta}~~~(ICCV'19)    & 15.2                                 & 20.3                                 & 29.0                                 & 25.5                                 & 28.7                                 & 13.4                                 & 20.6                                 & 28.6                                 & 32.4                                 & 38.8                                 & 19.6                                 & 20.8                                 & 28.7                                 & 42.2                                 & 42.1                                 \\
                         TFA-fc~\cite{wang2020frustratingly}~~~(ICML'20)    & 36.8                                 & 29.1                                 & 43.6                                 & 55.7 & 57.0                                 & 18.2                                 & 29.0                                 & 33.4                                 & 35.5                                 & 39.0                                 & 27.7                                 & 33.6                                 & 42.5                                 & 48.7                                 & 50.2 \\
                         TFA-cos~\cite{wang2020frustratingly}~~~(ICML'20)   &  39.8         & 36.1                                 & 44.7                                 & 55.7 & 56.0                                 & 23.5                                 & 26.9                                 & 34.1                                 & 35.1                                 & 39.1                                 & 30.8                                 & 34.8                                 & 42.8                                 & 49.5 & 49.8                                 \\
                         MPSR~\cite{wu2020multi}~~~(ECCV'20)            & 37.2                                 & 43.6                                 & 50.9                                 & 53.7                                 & 60.2                                 & 24.8                                 & 28.1                                 & 38.0                                 & 39.8 & 45.9                                 & 37.3                                 & 40.0 & 43.9 & 47.8                                 & 50.1 \\
                         Ret. R-CNN~\cite{fan2021generalized}~~~(CVPR'21)      & 42.4 & 45.8                                 & 45.9                                 & 53.7                                 & 56.1                                 & 21.7                                 & 27.8                                 & 35.2                                 & 37.0                                 & 40.3                                 & 30.2                                 & 37.6                                 & 43.0                                 & 49.7 & 50.1 \\
                         TFA+H~\cite{zhang2021hallucination}~~~(CVPR'21)         & 45.1 & 44.0                                 & 44.7                                & 55.0 & 55.9                                 & 23.2                                 & 27.5                                 & 35.1                                 & 34.9                                 & 39.0                                 & 30.5                                 & 35.1                                 & 41.4                                 & 49.0          & 49.3          \\
FSCE~\cite{sun2021fsce}~~~(CVPR'21)            & 37.6                                 & 44.7                                 & 46.9                                 & 52.2                                 & 60.3                                 & 24.5                                 & 30.1                                 & 38.2                                 & 40.4                                 & 45.9                                 & 25.4                                 & 34.2                                 & 42.3                                 & 48.7                                 & 50.3
\\
FADI~\cite{cao2021few}~~~(NeurIPS'21)            & 50.3	& 54.8	& 54.2	& 59.3	& 63.2	& 30.6& 	35.0	& 40.3& 	42.8& 	48.0	& 45.7& 	49.7& 	49.1& 	55.0	& 59.6
\\
LVC~\cite{kaul2022label}~~~(CVPR'22)            & 36.0	&40.1	&48.6	&57.0	&59.9	&22.3	&22.8	&39.2	&44.2	&47.8&	34.3&	43.4	&42.9&	52.0	&54.5
\\
LVC-PL~\cite{kaul2022label}~~~(CVPR'22)            & \textcolor{blue}{ \textbf{54.5}}	&53.2	&58.8	&63.2	&65.7	&\textcolor{blue}{ \textbf{32.8}}	&29.2	&\textcolor{blue}{ \textbf{50.7}}	&49.8	&50.6	&48.4	&52.7	&55.0	&59.6	&59.6
\\
DeFRCN~\cite{qiao2021defrcn}~~~(CVPR'21)            & 53.7	&\textcolor{blue}{ \textbf{59.5}}&\textcolor{blue}{ \textbf{61.2}}	&\textcolor{blue}{ \textbf{65.7}}	&\textcolor{blue}{ \textbf{66.6}}	&32.3	&\textcolor{blue}{ \textbf{42.0}}	&49.5	&\textcolor{blue}{ \textbf{52.4}}&	\textcolor{blue}{ \textbf{53.4}}&\textcolor{blue}{ \textbf{53.6}}	&\textcolor{blue}{ \textbf{56.2}}	&\textcolor{blue}{ \textbf{56.9}}	&\textcolor{blue}{ \textbf{61.9}}	&\textcolor{blue}{ \textbf{62.3}}

\\ \midrule
                         MPSR+Meta-Static            & 36.7                                 & 47.0 & 52.1 & 53.8                                 & 60.8 & 25.3 & 31.6 & 38.4 & 40.8 & 46.9 & 38.3 & 39.7                                 & 44.8 & 47.2                                 & 50.1 \\
                                                  MPSR+Meta-Dynamic            & 40.4                                & 47.5 & 51.9 &  54.9          & 60.5 & 25.6 & 31.7 & 38.5 & 40.6 & 46.7 & 37.6 & 40.2 & 44.7 & 49.1  & 50.3 \\
MPSR+Meta-ScaledDynamic           & 41.5                              & 47.9 & 52.7 &  55.4          & 60.9 & 25.7 & 32.2 & 38.9 & 40.8 & 46.8 & 38.5 & 40.9 & 45.9 & 49.0                                & 51.0\\
MPSR+Aug           & 39.5	& 47.1	& 53.2	& 54.9	& 59.5	& 26.2	& 31.0	& 39.7	& 41.8	& 47.8	& 38.0	& 37.8	& 45.2	& 48.4	& 50.9\\
MPSR+Meta-Static+Aug           & 40.9	& 47.6	& 53.6	& 54.7	& 60.2	& 26.5	& 31.6	& 38.9	& 42.2	& 47.3	& 38.7	& 38.1	& 45.8	& 48.2	& 50.8\\
MPSR+Meta-Dynamic+Aug           & 41.0	& 47.5	& 53.8	& 55.2	& 60.2	& 26.4	& 32.2	& 39.8	& 42.7	& 48.5	& 38.9	& 39.1	& 46.0	& 48.8	& 51.3\\
MPSR+Meta-ScaledDynamic+Aug           & 41.8	& 48.7	& 54.2	& 55.7	& 61.1	& 26.5	& 32.7	& 40.0	& 42.5	& 48.7	& 39.0	& 40.4	& 46.2	& 49.6	& 51.2\\

\midrule
DeFRCN+Meta-ScaledDynamic+Aug           & \textcolor{red}{ \textbf{58.4}}& \textcolor{red}{ \textbf{62.4}}	& \textcolor{red}{ \textbf{63.2}}	& \textcolor{red}{ \textbf{67.6}}	& \textcolor{red}{ \textbf{67.7}}	& \textcolor{red}{ \textbf{34.0}}	& \textcolor{red}{ \textbf{43.1	}}& \textcolor{red}{ \textbf{51.0}}	& \textcolor{red}{ \textbf{53.6}}	& \textcolor{red}{ \textbf{54.0}}	& \textcolor{red}{ \textbf{55.1}}	& \textcolor{red}{ \textbf{56.6}}	& \textcolor{red}{ \textbf{57.3}}	& \textcolor{red}{ \textbf{62.6}}& \textcolor{red}{ \textbf{63.7}}\\
\bottomrule
\end{tabular}%
}
    \caption{Comparison to fine-tuning based FSOD methods on the Pascal VOC dataset, with only novel classes. The best and the second-best results are marked with \textcolor{red}{red} and \textcolor{blue}{blue}. MPSR+Meta-Static, MPSR+Meta-Dynamic, and MPSR+Meta-ScaledDynamic represent meta-tuning results.}
    \label{pascal_novel_tl}
\end{scriptsize}
\end{center}
\end{table*}

%% file: pascal_gfsd_tl.tex
\begin{table*}
\begin{center}
\begin{scriptsize}
\resizebox{\textwidth}{!}{%
\begin{tabular}{c|ccccc|ccccc|ccccc}
\toprule
\multicolumn{1}{c|}{}                                & \multicolumn{5}{c|}{Split-1}                                                                                                                                                                     & \multicolumn{5}{c|}{Split-2}                                                                                                                                                                     & \multicolumn{5}{c}{Split-3}                                                                                                                                                                      \\
\multicolumn{1}{c|}{\multirow{-2}{*}{Method/Shot}} & 1                                    & 2                                    & 3                                    & 5                                    & 10                                   & 1                                    & 2                                    & 3                                    & 5                                    & 10                                   & 1                                    & 2                                    & 3                                    & 5                                    & 10                                   \\ \midrule
                               FRCN~\cite{yan2019meta}~~~(ICCV'19)                & 24.9                                 & 31.4                                 & 40.3                                 & 37.6                                 & 41.0                                 & 22.1                                 & 31.3                                 & 39.1                                 & 43.0                                 & 47.5                                 & 30.8                                 & 32.3                                 & 40.5                                 & 52.2                                 & 51.7                                 \\
                               TFA-fc~\cite{wang2020frustratingly}~~~(ICML'20)              & 50.4                                 & 42.6                                 & 56.2                                 & 65.4                                 & 66.1 & 29.7                                 & 42.4                                 & 47.0                                 & 49.0                                 & 52.1                                 & 41.3                                 & 47.4                                 & 55.6                                 & 60.6                                 & 61.6                                 \\
                               TFA-cos~\cite{wang2020frustratingly}~~~(ICML'20)             & 53.1 & 49.5                                 & 57.1                                 & 65.4 & 65.3                                 & 36.3                                 & 40.0                                 & 47.6                                 & 48.6                                 & 52.2                                 & 44.5                                 & 48.5                                 & 55.9 & 61.2 & 61.4 \\
                               MPSR~\cite{wu2020multi}~~~(ECCV'20)                & 45.8                                 & 52.5                                 & 59.3                                 & 61.8                                 & 65.5                                 & 36.0                                 & 39.7                                 & 49.8                                 & 51.7 & 56.9                                 & 47.6                                 & 49.9                                 & 54.5                                 & 58.1                                 & 60.0                                 \\
                               FSCE~\cite{sun2021fsce}~~~(CVPR'21)                & 50.7	&56.5	&58.1	&61.6	&66.1	&36.5	&42.4	&49.8	&51.5	&55.8	&38.2	&47.4	& 54.6	&59.9	&61.1                                 \\
                             Ret. R-CNN~\cite{fan2021generalized}~~~(CVPR'21)          & 55.6 & 58.5 & 58.6                                 & 64.5 & 66.2                                 & 34.3                                 & 41.5                                 & 49.2                                 & 51.0                                 & 54.0                                 & 44.1                                 & 51.6 & 56.4 & 61.9 & 62.2 \\ 
                             DeFRCN~\cite{qiao2021defrcn}~~~(CVPR'21)            & \textcolor{blue}{\textbf{63.3}}	& \textcolor{blue}{\textbf{67.3}}	& \textcolor{blue}{\textbf{68.1}}	& \textcolor{blue}{\textbf{71.1}}	& \textcolor{blue}{\textbf{71.2}}	& \textcolor{blue}{\textbf{45.9}}	& \textcolor{blue}{\textbf{54.7}}	& \textcolor{blue}{\textbf{60.3}}	& \textcolor{blue}{\textbf{62.8}}	& \textcolor{blue}{\textbf{63.1}}	& \textcolor{blue}{\textbf{63.7}}	& \textcolor{blue}{\textbf{65.4}}	& \textcolor{blue}{\textbf{65.5}}	& \textcolor{blue}{\textbf{68.8}}	& \textcolor{blue}{\textbf{69.2}} \\
                             
                             \midrule
 
                               MPSR+Meta-Static                 & 45.7                                 & 56.4                                 & 60.3 & 62.1                                 & 66.1 &36.7 & 43.7 & 50.3 & 52.7 & 57.9 & 48.6 & 51.2                                 & 55.5                                 & 57.8                                 & 60.1                                 \\
                               MPSR+Meta-Dynamic                & 50.2                                 & 57.2 & 60.6 & 63.3                                 & 67.0 & 37.0 & 43.9 & 50.4 & 52.5 & 57.8 & 47.9 & 51.8 & 55.4                                 & 59.1                                 & 60.2\\
                                MPSR+Meta-ScaledDynamic               & 51.0                                 & 57.3 & 60.9 & 63.3                                 & 67.1 & 37.1 & 44.1 & 50.7 & 52.5 & 57.7 & 48.7 & 52.1 & 56.1                                 & 59.0                                 & 60.5\\
                                MPSR+Aug              & 49.9	& 56.2	& 61.5	& 63.0	& 66.5	& 37.4	& 43.0	& 51.4	& 53.6	& 58.6	& 48.1	& 49.3	& 55.7	& 58.7	& 60.8\\
                                MPSR+Meta-Static+Aug              & 51.3	& 56.9	& 62.0	& 62.8	& 66.9	& 37.7	& 43.5	& 50.7	& 53.7	& 58.1	& 48.6	& 49.5	& 55.9	& 58.5	& 60.3\\
                                MPSR+Meta-Dynamic+Aug              & 51.3	& 56.8	& 62.1	& 63.3	& 67.0	& 37.8	& 44.2	& 51.7	& 54.3	& 59.3	& 48.9	& 50.5	& 56.5	& 59.0	& 61.2\\
                                MPSR+Meta-ScaledDynamic+Aug              & 51.9	& 57.6	& 62.4	& 63.7	& 67.6	& 37.8	& 44.9	& 51.9	& 54.2	& 59.4	& 49.2	& 51.9	& 56.7	& 59.7	& 61.1\\
\midrule
DeFRCN+Meta-ScaledDynamic+Aug           & \textcolor{red}{ \textbf{66.7}}& \textcolor{red}{ \textbf{69.3}}	& \textcolor{red}{ \textbf{69.8}}	& \textcolor{red}{ \textbf{72.2}}	& \textcolor{red}{ \textbf{72.1}}	& \textcolor{red}{ \textbf{47.7}}	& \textcolor{red}{ \textbf{55.8}}& \textcolor{red}{ \textbf{61.8}}	& \textcolor{red}{ \textbf{63.9}}	& \textcolor{red}{ \textbf{63.7}}	& \textcolor{red}{ \textbf{64.9}}	& \textcolor{red}{ \textbf{65.8}}	& \textcolor{red}{ \textbf{66.2}}	& \textcolor{red}{ \textbf{69.7}}& \textcolor{red}{ \textbf{70.2}}\\
\bottomrule

\end{tabular}%
}
\caption{Comparison to fine-tuning based G-FSOD methods on the Pascal VOC dataset, with both base and novel classes. The best and the second-best results are marked with \textcolor{red}{red} and \textcolor{blue}{blue}. The harmonic mean (HM) of the base and novel class mAPs is used for the calculation.}
\label{pascal_gfsd_tl}
\end{scriptsize}
\end{center}

\end{table*}

%% file: pascal_novel_ml.tex
\begin{table*}
\begin{center}
\begin{scriptsize}
\resizebox{\textwidth}{!}{%
\begin{tabular}{cc|ccccc|ccccc|ccccc}
\toprule
\multicolumn{2}{c|}{\multirow{2}{*}{Method/Shot}}                                          & \multicolumn{5}{c|}{Novel Set 1} & \multicolumn{5}{c|}{Novel Set 2} & \multicolumn{5}{c}{Novel Set 3}  \\
\multicolumn{2}{c|}{}                                                                        & 1    & 2    & 3    & 5    & 10   & 1    & 2    & 3    & 5    & 10   & 1    & 2    & 3    & 5    & 10   \\
\midrule
\multirow{12}{*}{\begin{tabular}[c]{@{}c@{}}ML\end{tabular}}     & M. R-CNN~\cite{yan2019meta}~~~(ICCV'19)   & 19.9 & 25.5 & 35.0 & 45.7 & 51.5 & 10.4 & 19.4 & 29.6 & 34.8 & 45.4 & 14.3 & 18.2 & 27.5 & 41.2 & 48.1 \\
                                                                              & M. R-CNN*~\cite{yan2019meta}~~~(ICCV'19)  & 16.8 & 20.1 & 20.3 & 38.2 & 43.7 & 7.7  & 12.0 & 14.9 & 21.9 & 31.1 & 9.2  & 13.9 & 26.2 & 29.2 & 36.2 \\
                                                                              & FSRW~\cite{kang2019few}~~~(ICCV'19)         & 14.8 & 15.5 & 26.7 & 33.9 & 47.2 & 15.7 & 15.3 & 22.7 & 30.1 & 39.2 & 19.2 & 21.7 & 25.7 & 40.6 & 41.3 \\
                                                                              & MetaDet~\cite{wang2019meta}~~~(ICCV'19)      & 18.9 & 20.6 & 30.2 & 36.8 & 49.6 & 21.8 & 23.1 & 27.8 & 31.7 & 43.0 & 20.6 & 23.9 & 29.4 & 43.9 & 44.1 \\
                                                                              & FsDet~\cite{xiao2020few}~~~(ECCV'20)   & 25.4 & 20.4 & 37.4 & 36.1 & 42.3 & 22.9 & 21.7 & 22.6 & 25.6 & 29.2 & 32.4 & 19.0 & 29.8 & 33.2 & 39.8 \\
                                                                              & TIP~\cite{li2021transformation}~~~(CVPR'21)   & 27.7 & 36.5 & 43.3 & 50.2 & 59.6 & 22.7 & 30.1 & 33.8 & 40.9 & 46.9 & 21.7 & 30.6 & 38.1 & 44.5 & 50.9 \\
                                                                             & DCNet~\cite{hu2021dense}~~~(CVPR'21)   & 33.9 & 37.4 & 43.7 & 51.1 & 59.6 & 23.2 & 24.8 & 30.6 & 36.7 & 46.6 & 32.3 & 34.9 & 39.7 & 42.6 & 50.7 \\
                                                                             & CME~\cite{li2021beyond}~~~(CVPR'21)   & 41.5 & 47.5& 50.4 & 58.2 & 60.9 & 27.2 & 30.2 & 41.4 & 42.5 & 46.8 & 34.3 & 39.6 & 45.1 & 48.3 & 51.5 \\
                                                                             & QA-FewDet~\cite{han2021query}~~~(ICCV'21)   & 41.0 & 33.2 & 35.3 & 47.5 & 52.0 & 23.5 & 29.4 & 37.9 & 35.9 & 37.1 & 33.2 & 29.4 & 37.6 & 39.8 & 41.5 \\
                                                                             & KFSOD~\cite{zhang2022kernelized}~~~(CVPR'22)   & 44.6 & - & 54.4 & 60.9 & 65.8 & \textcolor{red}{\textbf{37.8}} & - & 43.1 & 48.1 & 50.4 & 34.8 & - & 44.1 & 52.7 & 53.9\\
                                                                             & FCT~\cite{han2022few}~~~(CVPR'22)   & \textcolor{blue}{\textbf{49.9}} & 57.1 & 57.9 & \textcolor{blue}{\textbf{63.2}} & \textcolor{blue}{\textbf{67.1}} & 27.6 & 34.5 & \textcolor{blue}{\textbf{43.7 }}& \textcolor{blue}{\textbf{49.2}} & 51.2 & 39.5 & \textcolor{blue}{\textbf{54.7}} & 52.3 & 57.0 & 58.7
\\
& Meta-DETR~\cite{zhang2021meta}~~~(TPAMI'22)   & 40.6 & 51.4 & \textcolor{blue}{\textbf{58.0}} & 59.2 & 63.6 & \textcolor{blue}{ \textbf{37.0}}	 & \textcolor{blue}{ \textbf{36.6}}	 & \textcolor{blue}{ \textbf{43.7}}	 & 49.1 & \textcolor{red}{ \textbf{54.6}}	 & \textcolor{blue}{ \textbf{41.6}} & 45.9 & \textcolor{blue}{ \textbf{52.7}} & \textcolor{blue}{ \textbf{58.9}} & \textcolor{blue}{ \textbf{60.6}} \\
\midrule

\multirow{-1}{*}{Ours}     & DeFRCN+Meta-ScaledDynamic+Aug           & \textcolor{red}{ \textbf{58.4}}& \textcolor{red}{ \textbf{62.4}}	& \textcolor{red}{ \textbf{63.2}}	& \textcolor{red}{ \textbf{67.6}}	& \textcolor{red}{ \textbf{67.7}}	& 34.0& \textcolor{red}{ \textbf{43.1	}}& \textcolor{red}{ \textbf{51.0}}	& \textcolor{red}{ \textbf{53.6}}	& \textcolor{blue}{ \textbf{54.0}}	& \textcolor{red}{ \textbf{55.1}}	& \textcolor{red}{ \textbf{56.6}}	& \textcolor{red}{ \textbf{57.3}}	& \textcolor{red}{ \textbf{62.6}}& \textcolor{red}{ \textbf{63.7}}\\
\bottomrule

\end{tabular}%
}
\caption{Comparison to meta-learning based FSOD methods on the Pascal VOC dataset, with only novel classes. The best and the second-best results are marked with \textcolor{red}{red} and \textcolor{blue}{blue}. MPSR+Meta-Static, MPSR+Meta-Dynamic, and MPSR+Meta-ScaledDynamic represent meta-tuning results. ML represents the meta learning based methods.}
\label{pascal_novel_ml}
\end{scriptsize}
\end{center}
\end{table*}

%% file: pascal_gfsd_ml.tex
\begin{table*}
\begin{center}
\begin{scriptsize}
\resizebox{\textwidth}{!}{%
\begin{tabular}{cc|ccccc|ccccc|ccccc}
\toprule
\multicolumn{2}{c|}{}                                & \multicolumn{5}{c|}{Split-1}                                                                                                                                                                     & \multicolumn{5}{c|}{Split-2}                                                                                                                                                                     & \multicolumn{5}{c}{Split-3}                                                                                                                                                                      \\
\multicolumn{2}{c|}{\multirow{-2}{*}{Method/Shot}} & 1                                    & 2                                    & 3                                    & 5                                    & 10                                   & 1                                    & 2                                    & 3                                    & 5                                    & 10                                   & 1                                    & 2                                    & 3                                    & 5                                    & 10                                   \\  \midrule
                               & M. R-CNN~\cite{yan2019meta}~~~(ICCV'19)            & 17.3                                 & 25.3                                 & 27.3                                 & 44.4                                 & 50.4                                 & 11.6                                 & 18.5                                 & 21.9                                 & 30.8                                 & 41.3                                 & 13.3                                 & 20.2                                 & 33.4                                 & 38.0                                 & 45.5                                 \\
                               & FSRW~\cite{kang2019few}~~~(ICCV'19)                & 24.2                                 & 24.8                                 & 37.8                                 & 44.2                                 & 54.2                                 & 25.5                                 & 24.9                                 & 33.8                                 & 41.5                                 & 49.0                                 & 29.7                                 & 32.4                                 & 36.7                                 & 49.9                                 & 49.9                                 \\
\multirow{-3}{*}{ML}           & FsDet~\cite{xiao2020few}~~~(ECCV'20)               & 31.1                                 & 28.4                                 & 39.1                                 & 43.5                                 & 49.5                                 & 29.3                                 & 30.5                                 & 30.7                                 & 34.4                                 & 39.8                                 & 35.2                                 & 26.9                                 & 35.6                                 & 41.8                                 & 47.8                                 \\ \midrule

\multirow{-1}{*}{Ours}     & DeFRCN+Meta-ScaledDynamic+Aug           & \textcolor{red}{ \textbf{58.4}}& \textcolor{red}{ \textbf{62.4}}	& \textcolor{red}{ \textbf{63.2}}	& \textcolor{red}{ \textbf{67.6}}	& \textcolor{red}{ \textbf{67.7}}	& \textcolor{red}{ \textbf{34.0}}	& \textcolor{red}{ \textbf{43.1	}}& \textcolor{red}{ \textbf{51.0}}	& \textcolor{red}{ \textbf{53.6}}	& \textcolor{red}{ \textbf{54.0}}	& \textcolor{red}{ \textbf{55.1}}	& \textcolor{red}{ \textbf{56.6}}	& \textcolor{red}{ \textbf{57.3}}	& \textcolor{red}{ \textbf{62.6}}& \textcolor{red}{ \textbf{63.7}}\\
\bottomrule

\end{tabular}%
}
\caption{
Comparison to meta-learning based G-FSOD methods on the Pascal VOC dataset, with both base and novel classes. The best results are marked with \textcolor{red}{red}. The harmonic mean (HM) of the base and novel class mAPs is used for the calculation.}
\label{pascal_gfsd_ml}
\end{scriptsize}
\end{center}
\end{table*}

%% file: coco_novel_gfsd_ml.tex
\begin{table*}
\begin{center}
\begin{scriptsize}
\begin{tabular}{cc|cc|cc}
\toprule
\multicolumn{2}{c|}{}                              & \multicolumn{2}{c|}{Novel Classes}                                          & \multicolumn{2}{c}{All Classes (HM)}                                        \\
\multicolumn{2}{c|}{\multirow{-2}{*}{Method/Shot}} & 10-shot                              & 30-shot                              & 10-shot                              & 30-shot                              \\ \midrule
                              & ONCE~\cite{perez2020incremental}               & 1.2                                  & -                                    & 2.2                                  & -                                    \\
                              & Meta R-CNN~\cite{yan2019meta}         & 6.1                                  & 9.9                                  & 5.6                                  & 8.3                                  \\
                              & FSRW~\cite{kang2019few}               & 5.6                                  & 9.1                                  & -                                    & -                                    \\
                              & FsDetView~\cite{xian2017zero}          & 7.6                                  & 12.0                                 & 6.9                                  & 10.5                                 \\
                              & TIP~\cite{li2021transformation}                &  16.3 &  18.3 & -                                    & -                                    \\
                              & DCNET~\cite{duan2019centernet}              & 12.8                                 &  18.6 & -                                    & -                                    \\
                              & CME~\cite{li2021beyond}                &  15.1 & 16.9                                 & -                                    & -                                    \\
                              & QA-FewDet~\cite{han2021query}                & 10.2                                 & 11.5                                 & -                                    & - \\
                              & FCT~\cite{han2022few}                &  17.1 & 21.4                                 & -                                    & -                                    \\
                              & DAnA~\cite{chen2021dual}                &  18.6 & 21.6                                 & -                                    & -                                    \\
\multirow{-11}{*}{\begin{tabular}[c]{@{}c@{}}ML\end{tabular}}          
& Meta-DETR~\cite{zhang2021meta}                & \textcolor{red}{\textbf{19.0}}                                 & 22.2                                 & -                                    & - \\
\midrule
\multirow{-1}{*}{Ours}        & DeFRCN+Meta-ScaledDynamic+Aug           & 18.8                                & \textcolor{red}{\textbf{23.4}}                               &  \textcolor{red}{ \textbf{24.4}} &  \textcolor{red}{ \textbf{28.0}} \\
\bottomrule
\end{tabular}
\caption{FSOD and G-FSOD results on the MS COCO dataset with novel classes. The best results are marked with \textcolor{red}{red}. The harmonic mean (HM) of the base and novel class mAPs is used for the calculation.}
\label{coco_base_novel_ml}
\end{scriptsize}
\end{center}
\end{table*}

%% file: visual-results-appendix.tex
\begin{figure*}
\begin{center}
\resizebox{1\textwidth}{!}{%
\begin{tabular}{cccc}
\HangBox{\includegraphics[width=2.8cm, height=2.4cm]{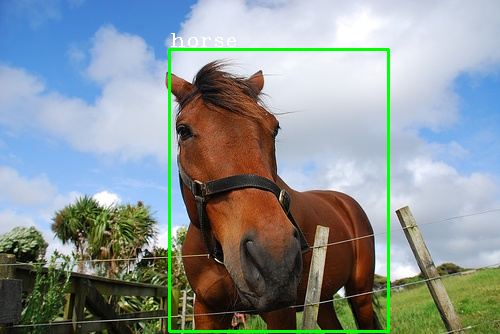}}
\HangBox{\includegraphics[width=2.8cm, height=2.4cm]{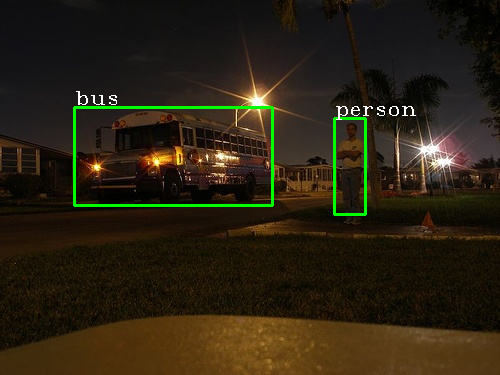}}
\HangBox{\includegraphics[width=2.8cm, height=2.4cm]{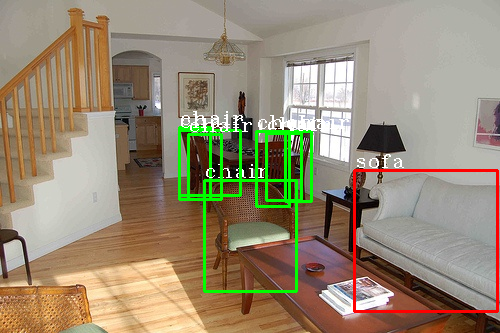}}
\HangBox{\includegraphics[width=2.8cm, height=2.4cm]{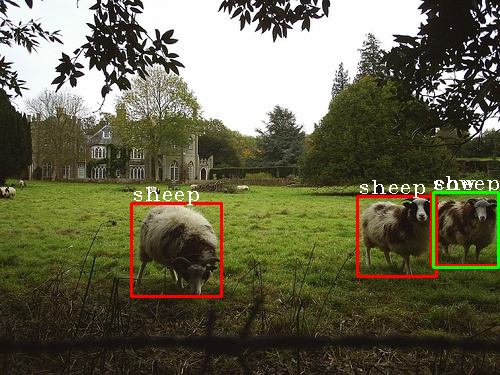}}
\\
\HangBox{\includegraphics[width=2.8cm, height=2.4cm]{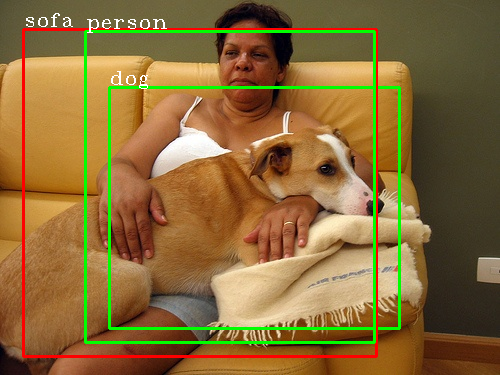}}
\HangBox{\includegraphics[width=2.8cm, height=2.4cm]{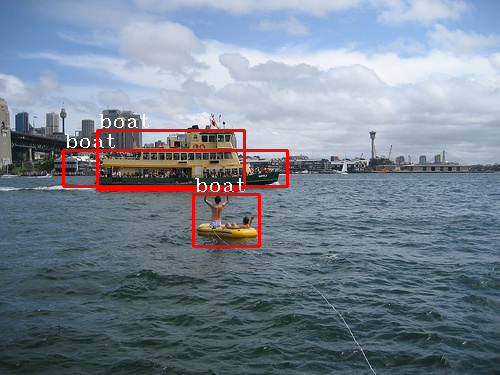}}
\HangBox{\includegraphics[width=2.8cm, height=2.4cm]{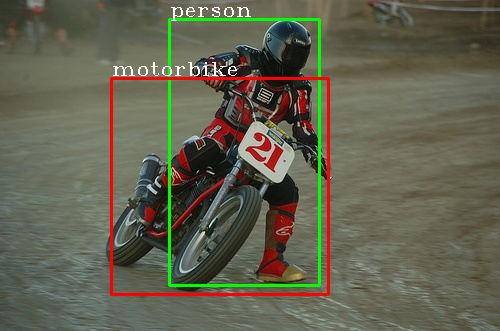}}
\HangBox{\includegraphics[width=2.8cm, height=2.4cm]{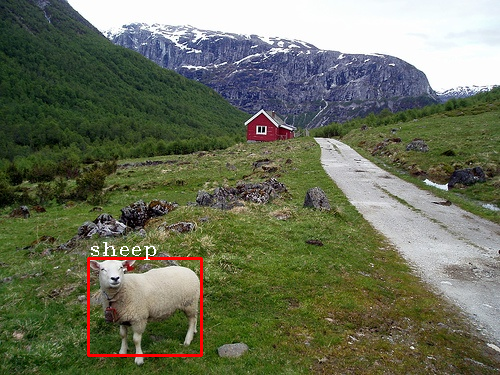}}
\\
\HangBox{\includegraphics[width=2.8cm, height=2.4cm]{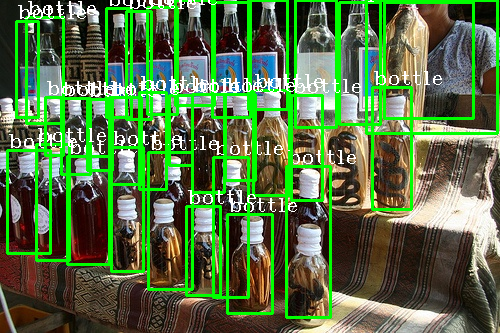}}
\HangBox{\includegraphics[width=2.8cm, height=2.4cm]{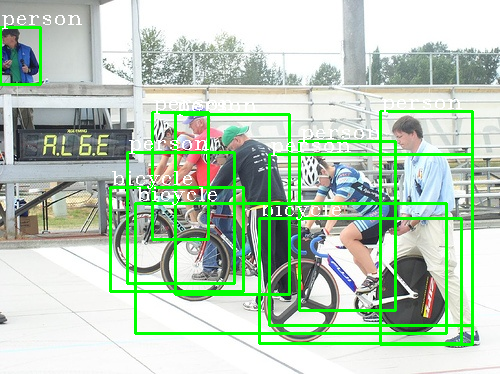}}
\HangBox{\includegraphics[width=2.8cm, height=2.4cm]{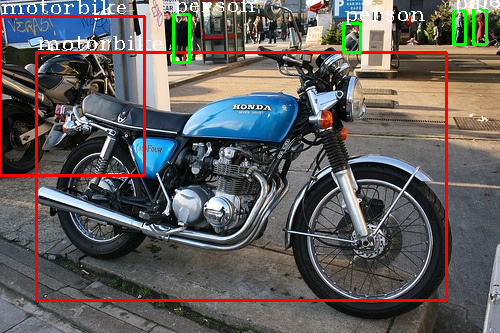}}
\HangBox{\includegraphics[width=2.8cm, height=2.4cm]{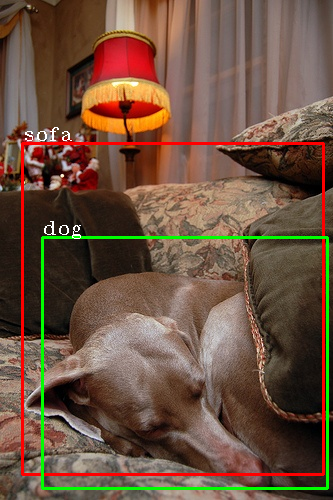}}
\\
\end{tabular}
}
\caption{Randomly sampled \textit{MPSR+Meta-ScaledDynamic+Aug} object detection results for the Pascal VOC dataset Split-3/10-shot experiment. Base class instance candidates are marked with \textcolor{green}{green}, and novel class instance candidates are marked with \textcolor{red}{red} color. (Best viewed in color.)}
\label{fig:visual_comparison_app}
\end{center}
\end{figure*}

%% file: visual-results-defrcn-appendix.tex
\begin{figure*}
\begin{center}
\resizebox{1\textwidth}{!}{%
\begin{tabular}{cccc}
\HangBox{\includegraphics[width=2.8cm, height=2.4cm]{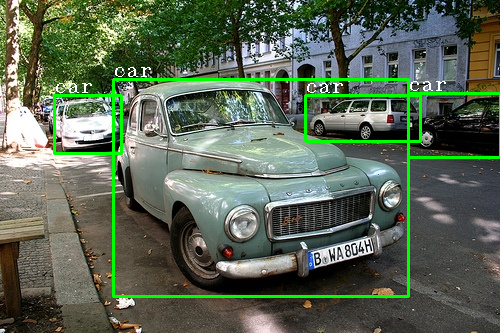}}
\HangBox{\includegraphics[width=2.8cm, height=2.4cm]{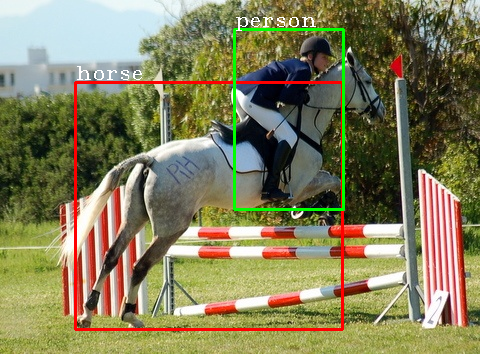}}
\HangBox{\includegraphics[width=2.8cm, height=2.4cm]{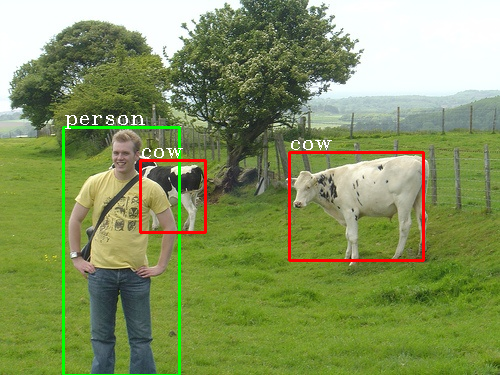}}
\HangBox{\includegraphics[width=2.8cm, height=2.4cm]{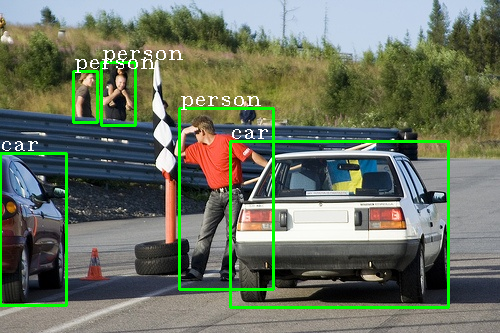}}
\\
\HangBox{\includegraphics[width=2.8cm, height=2.4cm]{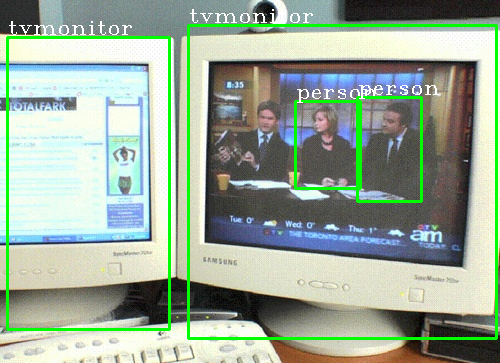}}
\HangBox{\includegraphics[width=2.8cm, height=2.4cm]{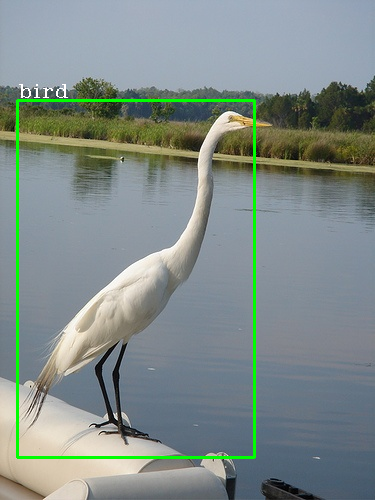}}
\HangBox{\includegraphics[width=2.8cm, height=2.4cm]{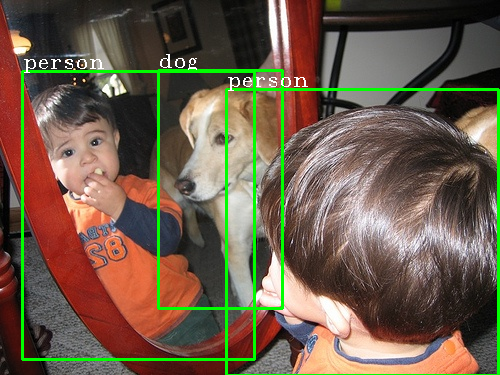}}
\HangBox{\includegraphics[width=2.8cm, height=2.4cm]{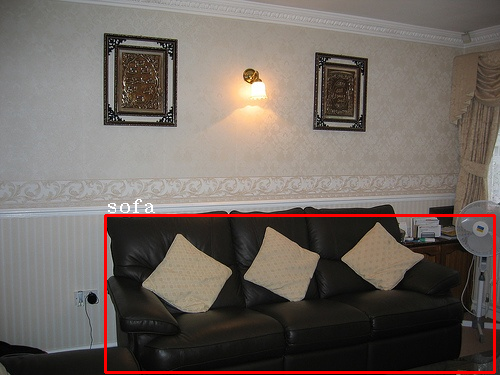}}
\\
\HangBox{\includegraphics[width=2.8cm, height=2.4cm]{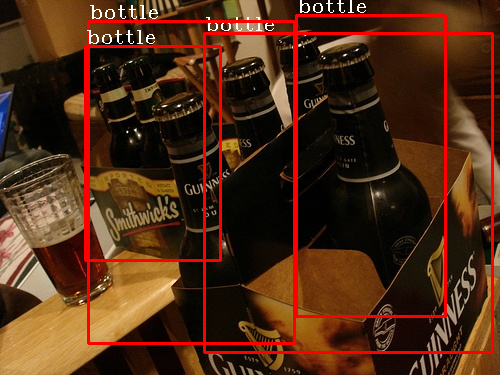}}
\HangBox{\includegraphics[width=2.8cm, height=2.4cm]{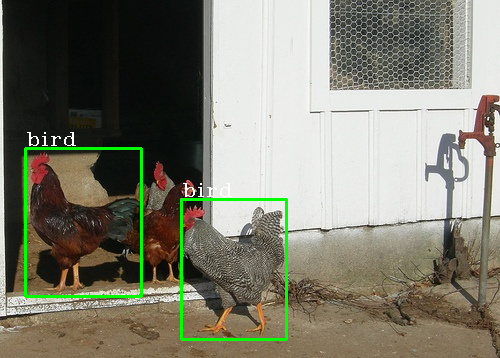}}
\HangBox{\includegraphics[width=2.8cm, height=2.4cm]{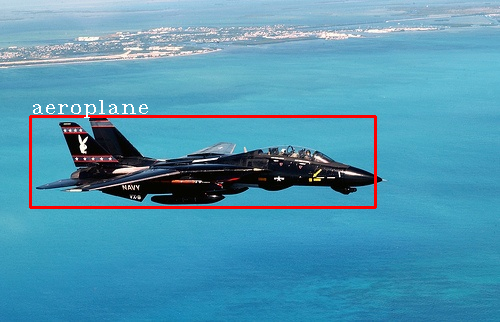}}
\HangBox{\includegraphics[width=2.8cm, height=2.4cm]{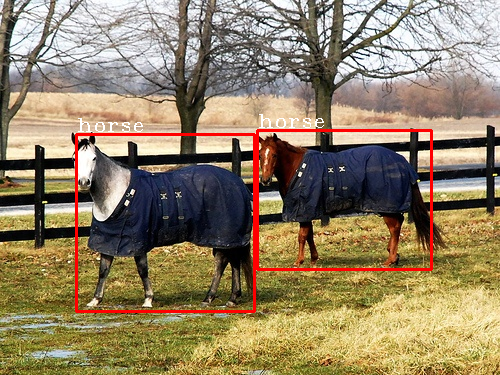}}
\\
\HangBox{\includegraphics[width=2.8cm, height=2.4cm]{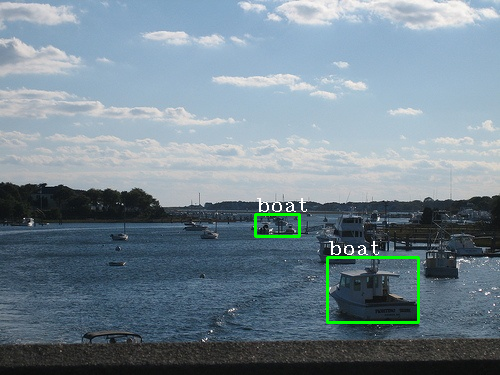}}
\HangBox{\includegraphics[width=2.8cm, height=2.4cm]{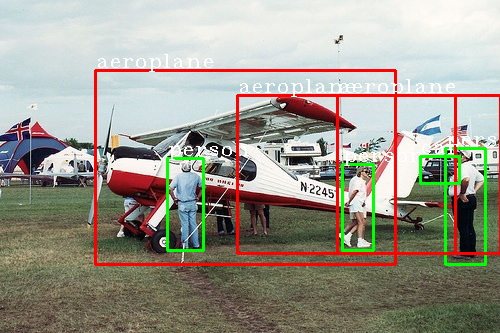}}
\HangBox{\includegraphics[width=2.8cm, height=2.4cm]{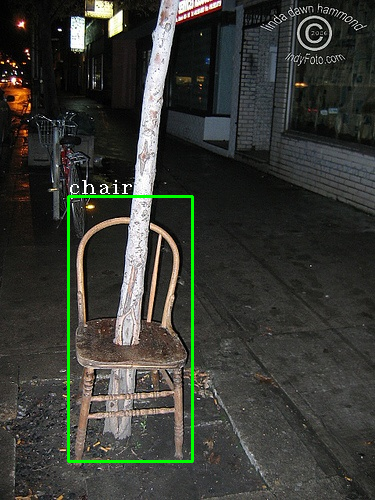}}
\HangBox{\includegraphics[width=2.8cm, height=2.4cm]{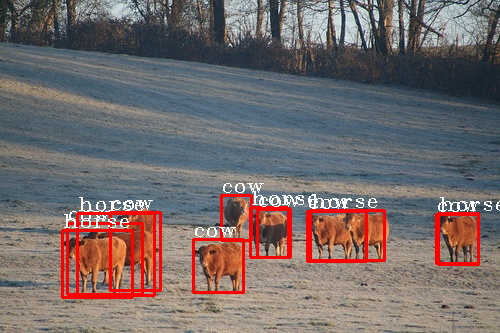}}
\\
\HangBox{\includegraphics[width=2.8cm, height=2.4cm]{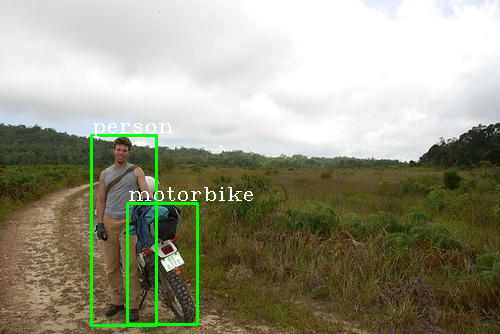}}
\HangBox{\includegraphics[width=2.8cm, height=2.4cm]{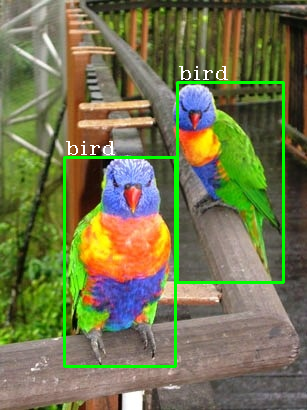}}
\HangBox{\includegraphics[width=2.8cm, height=2.4cm]{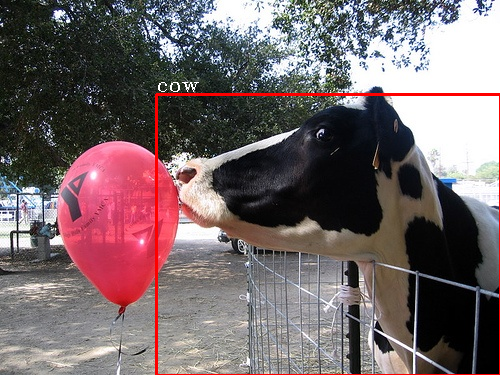}}
\HangBox{\includegraphics[width=2.8cm, height=2.4cm]{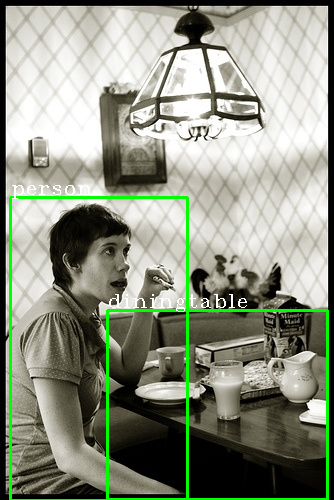}}
\\
\end{tabular}
}
\caption{Randomly sampled \textit{DeFRCN+Meta-ScaledDynamic+Aug} object detection results for the Pascal VOC dataset Split-2/10-shot experiment. Base class instance candidates are marked with \textcolor{green}{green}, and novel class instance candidates are marked with \textcolor{red}{red} color. (Best viewed in color.)}
\label{fig:visual_comparison_app_defrcn}
\end{center}
\end{figure*}